\documentclass[11pt]{article}

\usepackage[preprint]{acl}

\usepackage{times}
\usepackage{latexsym}

\usepackage[T1]{fontenc}

\usepackage[utf8]{inputenc}

\usepackage{microtype}

\usepackage{inconsolata}

\usepackage{graphicx}

\usepackage{comment}
\usepackage{amsmath}
\usepackage{booktabs}
\usepackage{multirow}
\usepackage[ruled,vlined,linesnumbered]{algorithm2e}
\usepackage{listings}
\usepackage{url}
\usepackage{siunitx}
\newcolumntype{P}{S[table-format=2.1] @{\,} >{\tiny}l}  
\usepackage[inkscapelatex=false]{svg}

\title{GRASP: Gated Regression-Aware Skill Proposer\\for Self-Improving LLM Agents}

\author{
  \textbf{Johannes Moll\textsuperscript{1}},
  \textbf{Jean-Philippe Corbeil\textsuperscript{2}},
  \textbf{Jiazhen Pan\textsuperscript{1}},
  \textbf{Martin Hadamitzky\textsuperscript{3}},
\\
  \textbf{Daniel Rueckert\textsuperscript{1}},
\textbf{Lisa Adams\textsuperscript{4,\dag}},
\textbf{Keno Bressem\textsuperscript{3,4,\dag}}
\\
 \small\textsuperscript{1}Technical University of Munich and TUM University Hospital,
  \small\textsuperscript{2}Microsoft Healthcare \& Life Sciences,\\
   \small\textsuperscript{3}Department of Cardiovascular Radiology and Nuclear Medicine, German Heart Center, TUM University Hospital,\\
   \small\textsuperscript{4}Department of Diagnostic and Interventional Radiology, Klinikum rechts der Isar, TUM University Hospital
   \\
  \small
    \textbf{Correspondence:} johannes.moll@tum.de \quad
    \textsuperscript{\dag}Equal senior authorship \\
    \small Project page: \url{https://jomoll.github.io/GRASP} \quad Code: \url{https://github.com/jomoll/GRASP}
  }

\begin{document}
\maketitle
\begin{abstract}
LLM agents acting in structured environments fail in operational rather than conversational ways, and reliability depends on procedural knowledge of the environment. Prior self-improvement methods accumulate natural-language guidance without checking that each new item preserves previously correct behavior, so a note that fixes one trajectory can silently regress another. We introduce \textsc{Grasp} (Gated Regression-Aware Skill Proposer), which treats agent improvement as a sequence of edits to a bounded skill library, admitting each candidate only if it produces a net improvement on a balanced held-out probe under a hard regression budget. We evaluate \textsc{Grasp} across five base models on two FHIR-based clinical benchmarks, which score procedural reliability against FHIR state rather than clinical correctness or patient outcomes. On MedAgentBench, \textsc{Grasp} lifts gpt-oss-120b from 40.6\% to 88.8\%, exceeds the strongest of five self-improvement baselines by 21.0 points, and improves every other base model by 17.2 to 40.3 points. Ablations attribute the gain to comparative proposal generation, the acceptance gate, and the hard regression budget rather than to skill writing itself, which without validation is no better than using no skills. Granting the same acceptance gate to all five baselines lifts each of them in-domain and none of them out of distribution, isolating the gain to the gate applied to a bounded, editable library rather than to held-out validation itself. The mechanism helps in non-clinical environments where tasks recur with verifiable structure and is flat where the action space is open-ended. Frozen libraries transfer across models and across benchmarks that share a tool-calling convention and degrade under interface mismatch.
\end{abstract}

\section{Introduction}
\label{sec:intro}
Structured environments such as electronic health records, databases, and web interfaces impose strict operational rules on LLM agents. Violations of these rules produce recurring errors that often go uncorrected. A clinical assistant working with an EHR, for instance, must retrieve the right resources, apply temporal eligibility criteria, avoid duplicate medication orders, and verify that changes were written~\citep{jiang2025medagentbench, moll2026agentic}. These failures reflect the structure of the environment rather than the particular case, so reliability depends on procedural knowledge of how to act in a specific environment, not on general language ability~\citep{hsiao2025procedural, roig2025llms}.

To acquire this knowledge, prior self-improvement methods accumulate natural-language guidance at inference, either as reflections on individual failures~\citep{shinn2023reflexion} or as rules extracted across many~\citep{zhao2024expel}, admitting each item as soon as it is produced (Appendix~\ref{app:related}). Without a check against existing behavior, a note that fixes one trajectory can silently regress another, and over time monotonic accumulation dilutes task-relevant content and degrades the agent's context~\citep{shi2023large, liu2024lost}. Avoiding both failures requires held-out validation of each candidate against previously correct behavior, together with a bounded library that supports modification and removal alongside addition.

\begin{figure*}[t]
\centering
\includegraphics[width=0.9\textwidth]{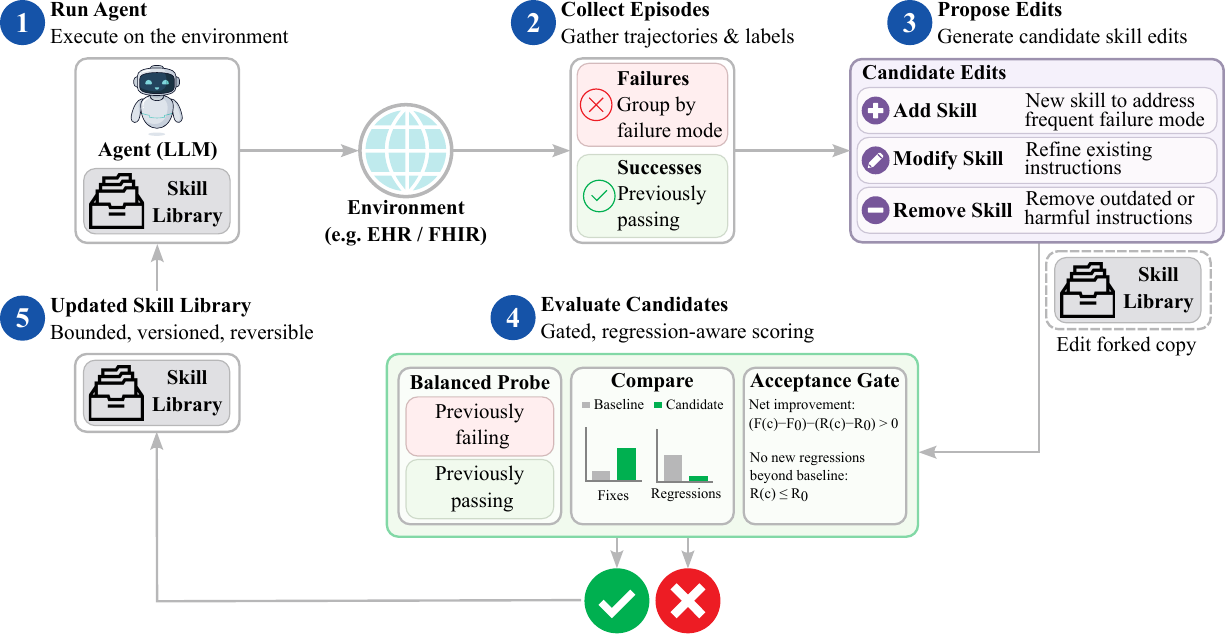}
\caption{Overview of \textsc{Grasp}. After each batch of development episodes, the agent's trajectories are partitioned into failures, grouped by mechanism-specific label, and previously passing episodes (1--2). A skill writer proposes $K$ candidate edits to the skill library, each adding, modifying, or removing a single skill (3). Every candidate is re-run on a balanced held-out probe of previously failing and previously passing episodes and scored against a fresh baseline of the current library. A candidate is applied only if it yields a net gain, $\bigl(F(c)-F_0\bigr)-\bigl(R(c)-R_0\bigr)>0$, and introduces no new regression, $R(c)\leq R_0$ (4). Rejected candidates leave the library unchanged. The accepted edit produces a new library version, which is injected into the agent's context for the next batch (5). The library stays capacity-bounded, versioned, and reversible, and no model parameters are updated.}
\label{fig:grasp}
\end{figure*}

We introduce \textsc{Grasp} (\textbf{G}ated \textbf{R}egression-\textbf{A}ware \textbf{S}kill \textbf{P}roposer), which treats agent self-improvement as a sequence of validated edits to a small library of behavioral instructions, or \textit{skills}, injected into the agent's context at inference~\citep{wang2023voyager}. After each batch of episodes, failed trajectories are grouped by failure mode, and candidate edits are proposed for the most frequent modes, each adding, modifying, or removing a skill. Each candidate is re-run on a balanced probe of previously failing and previously passing examples, and accepted only if it fixes more failures than it causes regressions, with no new regression beyond the existing baseline. The mechanism requires no parameter updates, but validating each candidate on a held-out probe dominates its training-time compute. The library remains small, versioned, and reversible. 

We evaluate \textsc{Grasp} across five language models (gpt-oss-120b, DeepSeek V4 Flash, Gemini 3.1 Flash Lite, GPT-4.1, GPT-5.4) on two FHIR-based clinical agent benchmarks, MedAgentBench~\citep{jiang2025medagentbench} and MedAgentBench-v2~\citep{chen2025medagentbench}, with supporting evaluation on FHIR-AgentBench~\citep{lee2025fhir} and four non-clinical environments~\citep{ALFWorld20, webshop2022, liu2024agentbench}. Our main results are as follows.
\begin{itemize}
    \item \textsc{Grasp} substantially outperforms a no-skills agent and five self-improvement baselines spanning memory, rule-extraction, and skill-library methods on MedAgentBench, an advantage that holds across all five base models.
    \item Ablations attribute the gain to comparative proposal generation, the acceptance gate, and the hard regression budget, showing that without them the skill-writing pipeline is no better than using no skills, which localizes the improvement to validated editing rather than skill writing.
    \item The mechanism generalizes beyond the clinical domain, helping in non-clinical environments where tasks recur with verifiable structure and remaining flat only where the action space is open-ended.
    \item Frozen skill libraries transfer across models and related benchmarks. A stronger writer improves weaker executors beyond what they learn for themselves while the reverse does not, an asymmetry that no ungated baseline reproduces, indicating that the libraries encode environment-specific procedural knowledge rather than model-specific patterns.
\end{itemize}

\section{\textsc{Grasp}: Gated Regression-Aware Skill Proposer}
\label{sec:method}

\textsc{Grasp} improves an LLM agent through repeated validated edits to a library of behavioral instructions, with no parameter updates. After each batch of training episodes, failed trajectories are grouped by failure mode, a skill-writing model proposes candidate edits for the most frequent modes, and each candidate is evaluated on a held-out probe before acceptance (Algorithm~\ref{alg:grasp}).

\subsection{Skills}
\label{sec:method:skills}

A \emph{skill} is a structured behavioral instruction injected into the agent's context at inference \citep{wang2023voyager}, specifying a trigger condition, a behavioral rule, an optional verification step for actions whose payload must satisfy a schema, and a contrastive example. Skills are Markdown documents with YAML frontmatter, making them auditable and reversible. The library starts empty and is populated by \textsc{Grasp} from training trajectories (Appendix~\ref{app:skill-template} and~\ref{app:injection}).

\subsection{Failure-Driven Skill Proposal}
\label{sec:method:proposal}
After each batch, a skill-writing model observes the failed traces and proposes edits, building on the verbal-feedback paradigm \citep{shinn2023reflexion, zhao2024expel} but operating on a structured library rather than a flat memory.

\paragraph{Mechanism-specific failure classification.} An LLM classifier assigns each failing trace an open-vocabulary label from the action sequence, the reported answer, and the expected answer. Labels must be mechanism-specific, meaning two labels imply two different corrective actions. A trace omitting a required date filter is labeled \texttt{date\_filter\_omitted} rather than \texttt{wrong\_answer}, since the former points to a corrective skill and the latter does not. Labels from prior epochs are presented as candidates, and new ones are minted when none fits (classifier dynamics in Appendix~\ref{app:classifier-dynamics}). Failures are grouped by label with no minimum group size, so a single failure can prompt a proposal if its mechanism is distinct.

\paragraph{Grouped proposal generation.} The framework generates $K$ proposals per update by cycling over failure-mode groups from largest to smallest. Each call to the skill writer provides the failing traces from one group, a sample of passing traces, active skill summaries with provenance and effectiveness statistics, and the labels of other active failure modes to discourage regressions on unrelated tasks. The skill writer returns an \texttt{ADD}, \texttt{MODIFY}, or \texttt{REMOVE} edit. To limit the number of skills, \texttt{ADD} is blocked at capacity unless a paired \texttt{REMOVE} frees a slot.

\subsection{Regression-Aware Selection}
\label{sec:method:selection}

Each candidate is evaluated against a balanced probe before being committed, following the held-out validation strategy familiar from prompt-optimization frameworks \citep{khattab2023dspy}. Probe examples are drawn exclusively from the development split, with validation and test examples never entering training (full per-batch loop in Appendix~\ref{app:algorithm}).

\paragraph{Probe construction.} The probe contains up to $N/2$ previously-failing and $N/2$ previously-passing samples from episodes completed earlier in the current epoch, stratified by task type, where \emph{previously-failing} and \emph{previously-passing} refer to per-sample correctness recorded under the library state in effect when each sample was originally run. For the first batch of each epoch, the previous epoch's samples serve as the baseline. The probe is always out-of-sample relative to the batch that generated the proposals.

\paragraph{Acceptance criterion.} Probe labels from earlier batches reflect library states that may differ from the current library $S$, so we re-run $S$ on the probe for a fresh causal baseline. Let $\mathcal{P}_{\mathrm{fail}}$ and $\mathcal{P}_{\mathrm{pass}}$ denote the previously-failing and previously-passing subsets. Running $S$ yields baseline counts $F_0 = |\{x \in \mathcal{P}_{\mathrm{fail}} : x \text{ passes under } S\}|$ and $R_0 = |\{x \in \mathcal{P}_{\mathrm{pass}} : x \text{ fails under } S\}|$. For a candidate library $S^c$, the analogous counts $F(c)$ and $R(c)$ are computed on the same probe. A candidate is accepted only if
\begin{equation}
  \bigl(F(c) - F_0\bigr) - \bigl(R(c) - R_0\bigr) > 0
  \quad \text{and} \quad
  R(c) \leq R_0.
  \label{eq:acceptance}
\end{equation}
The first condition requires a net improvement in fixes over regressions, the second is a hard regression budget. The highest-scoring candidate satisfying both is applied, otherwise the library is left unchanged. If the winning candidate introduces any regression, a contrastive revision step prompts the skill writer to narrow the trigger or add a guard clause, and the revision replaces the original only if its adjusted score is strictly higher and still satisfies the regression budget (Appendix~\ref{app:revision}).

\section{Experimental Setup}
\label{sec:setup}

\subsection{Benchmarks}
\label{sec:setup:benchmarks}

Our primary evaluation uses three FHIR-based clinical agent benchmarks. \textbf{MedAgentBench} \citep{jiang2025medagentbench} requires an agent to query a live FHIR server, reconcile medication orders, and construct write payloads that match an expected resource. \textbf{MedAgentBench-v2} \citep{chen2025medagentbench} extends it with multi-step decision tasks, coordinated writes, and clinical safety protocols. \textbf{FHIR-AgentBench} \citep{lee2025fhir} evaluates structured clinical QA and tool use on an independent FHIR environment. Each is split into disjoint development, validation, and held-out test sets (Table~\ref{tab:benchmarks}), with additional out-of-distribution (OOD) splits for the two MedAgentBench variants. We report exact-match accuracy against ground-truth answers and FHIR-server state (scoring details in Appendix~\ref{app:benchmarks}). For evidence of generality, we also evaluate on four AgentBench \citep{liu2024agentbench} environments, ALFWorld~\cite{ALFWorld20}, WebShop~\cite{webshop2022}, OS Interaction, and DBBench.

\subsection{Models}
\label{sec:setup:models}

We evaluate \textsc{Grasp} with five base models spanning open and proprietary families and a range of capability. The open-source models are gpt-oss-120b, an open model that fits on a single H200 GPU, and DeepSeek V4 Flash, a larger long-context alternative, both self-hosted. The proprietary models are Gemini 3.1 Flash Lite, GPT-5.4 (low reasoning effort), and GPT-4.1, three frontier models accessed via official provider APIs. Exact handles, sizes, precision settings, and access dates appear in Appendix~\ref{app:models}. Within a run, the same model serves as both executing agent and skill-writer.

\subsection{Baselines}
\label{sec:setup:baselines}

We compare \textsc{Grasp} against a no-skills baseline and five self-improvement methods, all using the same base agent, decoding settings, and training episodes. For every method, the same model serves as executing agent and as the LLM that writes notes, rules, or skills.

\paragraph{Sequential memory \citep{chen2025medagentbench}} appends a natural-language correction note to a flat memory block after each failing episode. Notes are task-specific conditional instructions, and the memory grows unbounded with no acceptance gate.

\paragraph{Batch memory} is identical but updated once per batch, matching \textsc{Grasp}'s cadence and isolating the acceptance gate from update frequency.

\paragraph{ExpeL \citep{zhao2024expel}} extracts contrastive natural-language rules by pairing each failing trace with its nearest successful trace and prompting an LLM to propose AGREE/EDIT/REMOVE/ADD operations on a bounded rule list. Rules are scored by accumulated operation counts and capped at 20.

\paragraph{Evo-MedAgent \citep{shen2026evo}} maintains episodic and semantic memory stores with top-$k$ retrieval and per-rule utility tracking, but no probe-based acceptance gate.

\paragraph{SkillX \citep{wang2026skillx}} extracts skills from successful trajectories via LLM decomposition and retrieves them by token overlap, with no regression control.

\noindent All methods inject into the same field of the task prompt, isolating the learned content and update rule from placement effects (Appendix~\ref{app:injection}).

\subsection{Training and Evaluation Protocol}
\label{sec:setup:protocol}

\paragraph{Training.} Each method trains for 5 epochs over the dev split with a batch size of 48 episodes. After every epoch, validation accuracy is computed, the learned state is checkpointed when it improves, and the best checkpoint is restored after the final epoch. The test and OOD splits are evaluated once on this checkpoint and never participate in training, checkpointing, or hyperparameter selection. For \textsc{Grasp}, the probe contains 36 examples from episodes completed earlier in the epoch, and the skill writer generates 4 candidate edits per batch.

\paragraph{Cross-model transfer.} For each transfer cell, the source model $M_s$ runs a complete \textsc{Grasp} training cycle as both executing agent and skill-writer. The resulting library is selected by best-validation on $M_s$, frozen, and applied to the target model $M_t$ at test time, with no further training, retrieval re-fitting, or library editing.

\begin{table*}[!h]
\centering
\caption{Main results on two FHIR clinical benchmarks. Val$^\star$ is the best validation accuracy across epochs and Test is the held-out test accuracy using the best-validation checkpoint, OOD is held-out task types. Open-source results are across five seeds, proprietary results across three. Bold marks the best score in each column within a model.}
\small
\begin{tabular}{ll PPP PPP}
\toprule
& & \multicolumn{6}{c}{MedAgentBench}
& \multicolumn{6}{c}{MedAgentBench-v2} \\
\cmidrule(lr){3-8}\cmidrule(lr){9-14}
Model & Method
& \multicolumn{2}{c}{Val$^\star$} & \multicolumn{2}{c}{Test} & \multicolumn{2}{c}{OOD}
& \multicolumn{2}{c}{Val$^\star$} & \multicolumn{2}{c}{Test} & \multicolumn{2}{c}{OOD} \\
\midrule
\multirow{7}{*}{gpt-oss-120b}
& No skills              & 36.4 & $\pm$5.6 & 40.6 & $\pm$3.9 & 8.7 & $\pm$3.2 & 57.6 & $\pm$6.5 & 61.1 & $\pm$4.6 & 75.2 & $\pm$7.3 \\
& Seq.\ Memory           & 38.5 & $\pm$12.4 & 41.2 & $\pm$5.5 & 16.7 & $\pm$15.6 & 54.2 & $\pm$10.2 & 53.8 & $\pm$11.6 & 82.3 & $\pm$4.5 \\
& Batch Memory           & 36.8 & $\pm$5.3 & 34.4 & $\pm$8.0 & 13.0 & $\pm$8.1 & 62.2 & $\pm$6.6 & 59.4 & $\pm$4.9 & 78.8 & $\pm$12.7 \\
& ExpeL                  & 47.5 & $\pm$6.2 & 49.1 & $\pm$6.3 & 12.0 & $\pm$5.2 & 65.5 & $\pm$1.9 & 67.8 & $\pm$2.8 & 86.3 & $\pm$1.4 \\
& Evo-MedAgent           & 62.5 & $\pm$17.0 & 67.8 & $\pm$14.1 & 31.3 & $\pm$21.7 & 65.3 & $\pm$3.5 & 66.9 & $\pm$2.6 & 86.7 & $\pm$0.0 \\
& SkillX                 & 51.5 & $\pm$4.1 & 53.1 & $\pm$4.6 & 21.3 & $\pm$3.6 & 67.5 & $\pm$3.7 & 64.1 & $\pm$4.2 & \bfseries 87.1 & \textbf{\tiny $\pm$0.8} \\
& \textsc{Grasp} (ours)  & \bfseries 86.0 & \textbf{\tiny $\pm$4.4} & \bfseries 88.8 & \textbf{\tiny $\pm$5.8} & \bfseries 56.3 & \textbf{\tiny $\pm$14.5} & \bfseries 77.0 & \textbf{\tiny $\pm$3.1} & \bfseries 76.9 & \textbf{\tiny $\pm$4.0} & 86.0 & $\pm$1.5 \\
\midrule
\multirow{7}{*}{DeepSeek V4 Flash}
& No skills              & 41.8 & $\pm$2.5 & 47.7 & $\pm$2.1 & 3.1 & $\pm$0.7 & 23.8 & $\pm$3.0 & 20.4 & $\pm$4.1 & 83.6 & $\pm$1.3 \\
& Seq.\ Memory           & 33.0 & $\pm$17.5 & 35.0 & $\pm$16.5 & 8.7 & $\pm$11.9 & 17.8 & $\pm$4.2 & 13.8 & $\pm$7.8 & 65.3 & $\pm$13.7 \\
& Batch Memory           & 36.5 & $\pm$6.7 & 37.8 & $\pm$6.4 & 3.3 & $\pm$4.7 & 17.8 & $\pm$5.6 & 16.6 & $\pm$5.3 & 57.3 & $\pm$28.9 \\
& ExpeL                  & 50.7 & $\pm$2.1 & 55.6 & $\pm$2.8 & 2.3 & $\pm$1.5 & 31.5 & $\pm$5.4 & 22.8 & $\pm$4.1 & \bfseries 85.0 & \textbf{\tiny $\pm$1.2} \\
& Evo-MedAgent           & 52.5 & $\pm$5.2 & 51.9 & $\pm$6.5 & 20.0 & $\pm$25.2 & 25.5 & $\pm$4.3 & 25.3 & $\pm$3.9 & 42.7 & $\pm$9.2 \\
& SkillX                 & 55.0 & $\pm$0.9 & 55.9 & $\pm$2.8 & 6.3 & $\pm$2.7 & 19.5 & $\pm$4.6 & 14.7 & $\pm$3.9 & 28.0 & $\pm$11.0 \\
& \textsc{Grasp} (ours)  & \bfseries 68.8 & \textbf{\tiny $\pm$3.2} & \bfseries 70.0 & \textbf{\tiny $\pm$5.5} & \bfseries 52.3 & \textbf{\tiny $\pm$14.7} & \bfseries 36.8 & \textbf{\tiny $\pm$13.5} & \bfseries 30.9 & \textbf{\tiny $\pm$14.7} & 67.0 & $\pm$22.5 \\
\midrule
\multirow{7}{*}{Gemini 3.1 Flash Lite}
& No skills              & 49.0 & $\pm$1.5 & 54.2 & $\pm$1.5 & 18.1 & $\pm$3.5 & 47.6 & $\pm$2.5 & 44.6 & $\pm$2.6 & 76.3 & $\pm$3.0 \\
& Seq.\ Memory           & 48.8 & $\pm$1.3 & 54.2 & $\pm$2.4 & 8.9 & $\pm$11.1 & 60.0 & $\pm$2.5 & 50.5 & $\pm$7.0 & 56.1 & $\pm$17.0 \\
& Batch Memory           & 29.2 & $\pm$13.8 & 39.6 & $\pm$22.2 & 8.3 & $\pm$10.1 & 55.4 & $\pm$4.7 & 51.6 & $\pm$5.6 & 73.9 & $\pm$3.8 \\
& ExpeL                  & 51.2 & $\pm$0.0 & 53.6 & $\pm$1.8 & 17.2 & $\pm$2.5 & 58.3 & $\pm$4.4 & 55.7 & $\pm$6.3 & 77.2 & $\pm$1.0 \\
& Evo-MedAgent           & 49.6 & $\pm$1.9 & 55.2 & $\pm$1.8 & 15.6 & $\pm$2.5 & 62.9 & $\pm$5.2 & 54.2 & $\pm$3.3 & 72.2 & $\pm$2.5 \\
& SkillX                 & 54.2 & $\pm$1.4 & 54.2 & $\pm$3.6 & 38.3 & $\pm$1.7 & 65.4 & $\pm$5.2 & 65.1 & $\pm$3.3 & 66.1 & $\pm$9.2 \\
& \textsc{Grasp} (ours)  & \bfseries 66.7 & \textbf{\tiny $\pm$2.6} & \bfseries 71.4 & \textbf{\tiny $\pm$1.8} & \bfseries 41.7 & \textbf{\tiny $\pm$20.9} & \bfseries 68.8 & \textbf{\tiny $\pm$5.7} & \bfseries 67.2 & \textbf{\tiny $\pm$2.7} & \bfseries 80.0 & \textbf{\tiny $\pm$3.3} \\
\midrule
\multirow{7}{*}{GPT-4.1}
& No skills              & 40.1 & $\pm$10.0 & 45.5 & $\pm$0.5 & 0.3 & $\pm$0.6 & 65.9 & $\pm$2.2 & 70.5 & $\pm$2.8 & \bfseries 84.9 & \textbf{\tiny $\pm$0.7} \\
& Seq.\ Memory           & 42.5 & $\pm$0.0 & 45.3 & $\pm$0.0 & 0.6 & $\pm$1.0 & 68.8 & $\pm$1.2 & 67.7 & $\pm$0.9 & 83.3 & $\pm$1.7 \\
& Batch Memory           & 46.2 & $\pm$8.8 & 47.4 & $\pm$9.5 & 2.2 & $\pm$2.5 & 68.3 & $\pm$5.1 & 73.4 & $\pm$1.6 & 83.9 & $\pm$1.0 \\
& ExpeL                  & 42.5 & $\pm$0.0 & 45.3 & $\pm$0.0 & 0.6 & $\pm$1.0 & 72.5 & $\pm$2.2 & 72.9 & $\pm$2.4 & 84.4 & $\pm$1.0 \\
& Evo-MedAgent           & 42.5 & $\pm$0.0 & 45.3 & $\pm$0.0 & 0.6 & $\pm$1.0 & 66.7 & $\pm$2.6 & 69.3 & $\pm$3.9 & 82.8 & $\pm$3.8 \\
& SkillX                 & 33.8 & $\pm$12.1 & 38.5 & $\pm$13.1 & 0.0 & $\pm$0.0 & 72.1 & $\pm$2.6 & \bfseries 74.0 & \textbf{\tiny $\pm$2.4} & 80.6 & $\pm$1.0 \\
& \textsc{Grasp} (ours)  & \bfseries 86.2 & \textbf{\tiny $\pm$2.5} & \bfseries 84.9 & \textbf{\tiny $\pm$0.9} & \bfseries 27.2 & \textbf{\tiny $\pm$16.9} & \bfseries 73.8 & \textbf{\tiny $\pm$3.3} & 70.3 & $\pm$1.6 & 82.8 & $\pm$1.0 \\
\midrule
\multirow{7}{*}{GPT-5.4 (low)}
& No skills              & 40.6 & $\pm$10.2 & 45.1 & $\pm$1.1 & 1.7 & $\pm$1.6 & 63.5 & $\pm$3.1 & 59.9 & $\pm$2.0 & 82.7 & $\pm$3.4 \\
& Seq.\ Memory           & 46.7 & $\pm$7.2 & 47.4 & $\pm$5.0 & 15.6 & $\pm$25.5 & 62.1 & $\pm$8.1 & 60.9 & $\pm$1.6 & 81.7 & $\pm$4.4 \\
& Batch Memory           & 47.9 & $\pm$0.7 & 46.4 & $\pm$0.9 & 16.7 & $\pm$3.3 & 62.9 & $\pm$11.2 & \bfseries 63.5 & \textbf{\tiny $\pm$1.8} & 78.3 & $\pm$10.9 \\
& ExpeL                  & 45.0 & $\pm$1.3 & 43.8 & $\pm$1.6 & 11.7 & $\pm$3.3 & 70.4 & $\pm$0.7 & 60.4 & $\pm$2.4 & 86.1 & $\pm$1.0 \\
& Evo-MedAgent           & 42.5 & $\pm$2.2 & 43.8 & $\pm$4.1 & 1.7 & $\pm$1.7 & 68.8 & $\pm$3.3 & 62.5 & $\pm$1.6 & 80.0 & $\pm$1.7 \\
& SkillX                 & 45.4 & $\pm$3.1 & 44.8 & $\pm$0.9 & 3.3 & $\pm$2.9 & 68.8 & $\pm$1.2 & 57.8 & $\pm$1.6 & \bfseries 86.7 & \textbf{\tiny $\pm$1.7} \\
& \textsc{Grasp} (ours)  & \bfseries 84.6 & \textbf{\tiny $\pm$10.5} & \bfseries 85.4 & \textbf{\tiny $\pm$10.4} & \bfseries 80.6 & \textbf{\tiny $\pm$6.7} & \bfseries 71.2 & \textbf{\tiny $\pm$1.8} & 63.3 & $\pm$3.3 & 80.0 & $\pm$4.7 \\
\bottomrule
\end{tabular}
\label{tab:main}
\end{table*}

\paragraph{Compute and context budget.} Methods differ in where they spend training compute and how much context they inject at inference. \textsc{Grasp} uses roughly $3.7\times$ more training-time LLM calls per batch than the simplest memory baselines, almost entirely from probe-based validation, since each batch evaluates the current library plus the candidates the skill-writer returns, at most $K$ and 3.68 on average, on a 36-episode probe (411 probe calls per batch, Appendix Table~\ref{tab:budget}). Section~\ref{sec:results:gated} reports the same accounting for the gated baselines and shows that this cost does not explain the gain. At inference the picture inverts. \textsc{Grasp} libraries average 5 skills and 5.6k tokens, while the memory and skill-library baselines grow to 34--53k (Appendix Table~\ref{tab:budget}). Only \textsc{Grasp} and ExpeL stay below 10k inference tokens.

\paragraph{Reporting and statistics.} Headline numbers are test accuracy unless otherwise indicated, reported as mean $\pm$ standard deviation across seeds. Main results and benchmark transfer use five seeds for open-source models and three for proprietary, while ablations, sensitivity analyses, and model transfer experiments use three seeds. For every major claim in Section~\ref{sec:results}, Appendix~\ref{app:significance} reports the mean difference $\Delta$ with a 95\% bootstrap CI from 10{,}000 resamples, Cohen's $d$, and permutation $p$-values.

\subsection{Ablations}
\label{sec:setup:ablations}
We ablate eight design choices of \textsc{Grasp} on MedAgentBench with gpt-oss-120b across three seeds. \emph{No failure grouping} passes all failed traces to the skill-writer at once rather than grouped by mechanism. \emph{No regression budget} accepts edits on positive net fixes without enforcing $R(c) \le R_0$. \emph{Fixes-only} scores candidates on previously failing examples alone, without the passing examples that detect regressions. \emph{Append-only} restricts the skill-writer to \texttt{ADD}. Two \emph{no acceptance gate} variants apply every edit unconditionally with FIFO eviction, at $K{=}4$ and $K{=}1$. Two \emph{matched-compute} variants separate the probe's validation signal from the cost of running it, scoring every candidate on the full probe as \textsc{Grasp} does but selecting without the scores, by the skill-writer's preference or at random. Sweeps over $B$, $N$, $K$, and the invalid-action weight $\lambda$ appear in Appendix~\ref{app:sensitivity}.

\section{Results}
\label{sec:results}
\subsection{Benchmark Results}
\label{sec:results:clinical}
Table~\ref{tab:main} reports test accuracy on MedAgentBench and MedAgentBench-v2. On MedAgentBench, \textsc{Grasp} is the strongest method on every model and on both the in-domain and OOD splits. In-domain, it lifts gpt-oss-120b from 40.6\% to 88.8\%, a gain of 48.2 points over the no-skills baseline and 21.0 over the strongest baseline (Evo-MedAgent, 67.8\%), and the pattern replicates across families, with gains of 17.2 to 40.3 points on the other four models. On the OOD split the lead is larger, reaching 56.3\% on gpt-oss-120b against 31.3\% for the best baseline, with margins above 25 points on DeepSeek, GPT-4.1, and GPT-5.4, and a narrower lead on Gemini (41.7\% against 38.3\%). On MedAgentBench-v2 the picture is more mixed. \textsc{Grasp} leads the in-domain test split on three of the five models, improving gpt-oss-120b from 61.1\% to 76.9\% (+15.8) and exceeding the best baseline (ExpeL, 67.8\%) by 9.1 points, with smaller leads on DeepSeek (30.9\% vs 25.3\%) and Gemini (67.2\% vs 65.1\%). On GPT-4.1 and GPT-5.4 it is within noise of the strongest baseline, where the dominant failure is exhausting the 8-action budget on two paginated-search task types and no method beats no-skills (Appendix~\ref{app:budget-failures}). On the OOD split the methods are closely matched and \textsc{Grasp} does not lead overall, scoring 86.0\% on gpt-oss-120b against 87.1\% for the best baseline (SkillX) with overlapping standard deviations. We make no superiority claim on v2 OOD.

\subsection{Learning Stability}
\label{sec:results:stability}

Figure~\ref{fig:learning_curves} shows validation accuracy across training epochs on MedAgentBench for gpt-oss-120b. \textsc{Grasp} improves almost monotonically and stabilizes above 85\% from epoch~3. Sequential and Batch memory regress below the no-skills baseline as memory accumulates, ExpeL and SkillX stagnate near baseline, and Evo-MedAgent reaches competitive validation accuracy on some epochs but oscillates by more than 20 points between adjacent checkpoints. Seed variance corroborates the dynamic, with Evo's standard deviation across five seeds of 14.1 against \textsc{Grasp}'s 5.8 at a 21.0-point higher mean.

\subsection{Cross-Model Transfer}
\label{sec:results:cross-model}

Table~\ref{tab:cross-model-transfer} reports cross-model transfer on MedAgentBench. Each column fixes an executor model and each row identifies the source model that produced the frozen skill library applied at inference. Diagonal entries reproduce the in-domain results of Table~\ref{tab:main}, and off-diagonal entries apply a library to a different executor without further adaptation. The clearest effect is on OOD task types, where libraries written by GPT-5.4 (low) improve every executor over its own self-trained library. Applied to gpt-oss-120b, GPT-5.4 skills reach 77.8\% OOD against 56.3\% from gpt-oss-120b's own library, a 21.5-point gain, while trading 12.8 points in-domain (88.8\% to 76.0\%). Applied to Gemini 3.1 Flash Lite, they lift OOD from 41.7\% to 71.1\% (+29.4) and in-domain test from 71.4\% to 76.6\% (+5.2). GPT-5.4 is the strongest cross-model source on every transfer cell except Gemini in-domain test, where the gpt-oss library is marginally stronger. Libraries from weaker source models applied to stronger executors are consistently worse than self-training. Off-diagonal transfer with Gemini as the source is high-variance (OOD standard deviations above 30 points) and we draw no conclusions from those cells.

\begin{table}[!h]
    \centering
    \caption{Cross-model skill transfer on MedAgentBench across three seeds. Rows indicate the skill library's
    training model; columns indicate the executor at evaluation time. Diagonal entries (source equals executor)
    reproduce the \textsc{Grasp} rows of Table~\ref{tab:main}; off-diagonal entries apply frozen skills without
    further training.}
    \small
    \setlength{\tabcolsep}{4pt}
    \begin{tabular}{@{}ll PPP@{}}
    \toprule
    & & \multicolumn{6}{c}{\textbf{Executor model}} \\
    \cmidrule(lr){3-8}
    \textbf{Skill source $\downarrow$} & \textbf{Split}
    & \multicolumn{2}{c}{\textbf{gpt-oss}}
    & \multicolumn{2}{c}{\textbf{Gemini 3.1}}
    & \multicolumn{2}{c}{\textbf{GPT-5.4}} \\
    \midrule
    \multirow{2}{*}{None}
      & Test & 40.6 & $\pm$3.9 & 54.2 & $\pm$1.5 & 45.1 & $\pm$1.1 \\
      & OOD  &  8.7 & $\pm$3.2 & 18.1 & $\pm$3.5 &  1.7 & $\pm$1.6 \\
    \midrule
    \multirow{2}{*}{gpt-oss}
      & Test & \bfseries 88.8 & \textbf{\tiny $\pm$5.8} & \bfseries 79.7 & \textbf{\tiny $\pm$7.8} & 72.4 & $\pm$15.8 \\
      & OOD  & 56.3 & $\pm$14.5 & 36.1 & $\pm$14.0 & 51.7 & $\pm$2.9 \\
    \midrule
    \multirow{2}{*}{Gemini 3.1}
      & Test & 65.6 & $\pm$8.3 & 71.4 & $\pm$1.8 & 69.3 & $\pm$6.5 \\
      & OOD  & 57.8 & $\pm$31.9 & 41.7 & $\pm$20.9 & 58.9 & $\pm$37.5 \\
    \midrule
    \multirow{2}{*}{GPT-5.4}
      & Test & 76.0 & $\pm$7.7 & 76.6 & $\pm$10.9 & \bfseries 85.4 & \textbf{\tiny $\pm$10.4} \\
      & OOD  & \bfseries 77.8 & \textbf{\tiny $\pm$1.0} & \bfseries 71.1 & \textbf{\tiny $\pm$1.9} & \bfseries 80.6 & \textbf{\tiny $\pm$6.7} \\
    \bottomrule
    \end{tabular}
    \label{tab:cross-model-transfer}
    \end{table}
\begin{table}[!h]
\centering
\caption{Cross-benchmark skill transfer on gpt-oss-120b. Rows indicate the skill library's training benchmark, and columns indicate the evaluation target. Diagonal entries (source equals target) reproduce the gpt-oss-120b row of Table~\ref{tab:main}, off-diagonal entries apply frozen skills without further training. DBBench (AgentBench) is a non-clinical control and FHIR-AB has no OOD split.}
\small
\setlength{\tabcolsep}{4pt}
\begin{tabular}{@{}ll PPP@{}}
\toprule
& & \multicolumn{6}{c}{\textbf{Target benchmark}} \\
\cmidrule(lr){3-8}
\textbf{Skill source $\downarrow$} & \textbf{Split}
& \multicolumn{2}{c}{\textbf{MAB}}
& \multicolumn{2}{c}{\textbf{MAB-v2}}
& \multicolumn{2}{c}{\textbf{FHIR-AB}} \\
\midrule
\multirow{2}{*}{None (baseline)}
  & Test & 40.6 & $\pm$3.9 & 61.1 & $\pm$4.6 & 51.5 & $\pm$2.2 \\
  & OOD  &  8.7 & $\pm$3.2 & 75.2 & $\pm$7.3 & \multicolumn{2}{c}{---} \\
\midrule
\multirow{2}{*}{MAB}
  & Test & \bfseries 88.8 & \textbf{\tiny $\pm$5.8} & \bfseries 78.1 & \textbf{\tiny $\pm$7.8} & 47.2 & $\pm$6.7 \\
  & OOD  & \bfseries 56.3 & \textbf{\tiny $\pm$14.5} & \bfseries 86.7 & \textbf{\tiny $\pm$0.0} & \multicolumn{2}{c}{---} \\
\midrule
\multirow{2}{*}{MAB-v2}
  & Test & 51.6 & $\pm$6.8 & 76.9 & $\pm$4.0 & 45.7 & $\pm$9.4 \\
  & OOD  & 34.4 & $\pm$22.8 & 86.0 & $\pm$1.5 & \multicolumn{2}{c}{---} \\
\midrule
\multirow{2}{*}{FHIR-AB}
  & Test & 45.3 & $\pm$4.4 & 67.2 & $\pm$8.8 & \bfseries 58.2 & \textbf{\tiny $\pm$3.4} \\
  & OOD  & 15.8 & $\pm$13.0 & 81.7 & $\pm$0.0 & \multicolumn{2}{c}{---} \\
\midrule
\multirow{2}{*}{DBBench}
  & Test & 38.5 & $\pm$2.4 & 57.8 & $\pm$7.2 & 48.3 & $\pm$3.1 \\
  & OOD  &  1.1 & $\pm$1.0 & 75.0 & $\pm$10.4 & \multicolumn{2}{c}{---} \\
\bottomrule
\end{tabular}
\label{tab:cross-benchmark-transfer}
\end{table}

\subsection{Cross-Benchmark Transfer}
\label{sec:results:cross-benchmark}

Table~\ref{tab:cross-benchmark-transfer} reports cross-benchmark transfer on gpt-oss-120b. Each row fixes a target benchmark and split, and each column applies a frozen library trained on a different source. Libraries trained on the DBBench environment~\cite{liu2024agentbench} serve as a non-clinical control.

Transfer between MedAgentBench and MedAgentBench-v2 is positive in both directions. MedAgentBench skills applied to MedAgentBench-v2 reach 78.1\% test and 86.7\% OOD, exceeding v2 self-training (76.9\% and 86.0\%). In reverse, v2 skills applied to MedAgentBench reach 51.6\% test against a 40.6\% baseline, with positive but high-variance OOD transfer (34.4\% against an 8.7\% baseline, $\pm$22.8). On FHIR-AgentBench, self-training improves gpt-oss-120b from 51.5\% to 58.2\% across five seeds ($\Delta = +6.8$, 95\% CI $[3.7, 10.0]$), while every library transferred into it degrades the baseline (MedAgentBench 47.2\%, MedAgentBench-v2 45.7\%, DBBench 48.3\%). Transfer in the reverse direction, out of FHIR-AgentBench onto the MedAgentBench family, does not show this degradation. The DBBench control transfers negatively or neutrally to every clinical target.

\subsection{Ablations}
\label{sec:results:ablations}
Table~\ref{tab:ablation} ablates the \textsc{Grasp} design choices defined in Section~\ref{sec:setup:ablations} on MedAgentBench with gpt-oss-120b. The full method reaches 88.8\% test accuracy. Removing any single component while keeping the acceptance gate stays above 80\%, above every comparator in Table~\ref{tab:main}. Removing failure grouping drops to 84.4\% ($-$4.4), the regression budget to 81.8\% ($-$7.0), and fixes-only and append-only both reach 80.2\% ($-$8.6), with fixes-only the least stable at $\pm$13.7.
\begin{table}[!t]
\centering
\caption{Ablation study on MedAgentBench with gpt-oss-120b across three seeds. Variants are defined in Section~\ref{sec:setup:ablations}. The \emph{matched compute} rows run the full probe as \textsc{Grasp} does but select without its scores.}
\small
\begin{tabular}{l PP}
\toprule
Method & \multicolumn{2}{c}{Val$^\star$} & \multicolumn{2}{c}{Test} \\
\midrule
\textsc{Grasp} (full)              & 86.0 & $\pm$4.4 & 88.8 & $\pm$5.8 \\
\midrule
\quad w/o failure grouping         & 82.1 & $\pm$5.2 & 84.4 & $\pm$3.1 \\
\quad w/o regression budget        & 81.2 & $\pm$6.5 & 81.8 & $\pm$1.8 \\
\quad fixes-only selection         & 78.8 & $\pm$14.4 & 80.2 & $\pm$13.7 \\
\quad append only                  & 73.3 & $\pm$8.3 & 80.2 & $\pm$9.4 \\
\midrule
\quad w/o acceptance gate ($K{=}4$) & 65.4 & $\pm$9.7 & 63.5 & $\pm$3.9 \\
\quad w/o acceptance gate ($K{=}1$) & 38.8 & $\pm$8.2 & 40.1 & $\pm$11.3 \\
\midrule
\quad matched compute (proposer) & 67.1 & $\pm$14.8 & 70.8 & $\pm$14.0 \\
\quad matched compute (random)   & 62.1 & $\pm$7.5  & 67.2 & $\pm$10.2 \\
\bottomrule
\end{tabular}
\label{tab:ablation}
\end{table}

Removing the acceptance gate entirely is more damaging. Applying every proposed edit unconditionally falls to 63.5\% ($-$25.3) at $K{=}4$ and to 40.1\% ($-$48.7) at $K{=}1$, the latter matching the no-skills baseline (40.6\%) within seed noise.
The matched-compute variants spend \textsc{Grasp}'s full per-batch probe budget but select the applied edit without the probe scores, reaching 70.8\% (proposer) and 67.2\% (random), within the variance of the no-gate $K{=}4$ variant (63.5\%).

\subsection{Granting the Gate to the Baselines}
\label{sec:results:gated}
 
The ablations isolate the gate within \textsc{Grasp} but leave open whether any method benefits equally from held-out validation. We therefore added the same gate to all five baselines on MedAgentBench with gpt-oss-120b across five seeds, sharing the gate implementation and varying only the proposal, application, and forking steps; \textsc{Grasp} re-run through it reproduces Table~\ref{tab:main}. \emph{Gate only} fires the gate at each method's native proposal count ($K{=}1$), and \emph{gate and comparative generation} ($K{=}4$) gives each baseline \textsc{Grasp}'s full configuration, that is everything \textsc{Grasp} has except a bounded, editable library. At $K{=}1$ every baseline improves in-domain, by $+1.6$ to $+15.0$ points, and none improves out of distribution, where the mean change is $-0.1$ and no individual change exceeds seed noise, while \textsc{Grasp} under the identical gate gains $+33.3$ and $+23.7$ (Table~\ref{tab:gated}). At $K{=}4$ no gated baseline exceeds 62.2 test or 26.1 OOD, against 88.8 and 56.3.

\begin{table*}[!h]
\centering
\caption{Granting the acceptance gate to all five baselines on MedAgentBench with gpt-oss-120b across five seeds. \emph{Ungated} reproduces Table~\ref{tab:main}, except for \textsc{Grasp}, whose ungated row is the no-gate $K{=}1$ ablation of Table~\ref{tab:ablation}. \emph{Gate only} adds the acceptance gate at each method's native proposal count ($K{=}1$). \emph{Gate + comp.} gives each method \textsc{Grasp}'s full configuration ($K{=}4$). Probe calls are per training batch under the $K{=}4$ configuration.}
\small
\setlength{\tabcolsep}{4pt}
\begin{tabular}{l PP PP PP r}
\toprule
& \multicolumn{4}{c}{\textbf{Ungated}} & \multicolumn{4}{c}{\textbf{Gate only} ($K{=}1$)} & \multicolumn{4}{c}{\textbf{Gate + comp.} ($K{=}4$)} & \\
\cmidrule(lr){2-5}\cmidrule(lr){6-9}\cmidrule(lr){10-13}
\textbf{Method}
& \multicolumn{2}{c}{Test} & \multicolumn{2}{c}{OOD}
& \multicolumn{2}{c}{Test} & \multicolumn{2}{c}{OOD}
& \multicolumn{2}{c}{Test} & \multicolumn{2}{c}{OOD}
& \textbf{Probe} \\
\midrule
No skills       & 40.6 & $\pm$3.9 & 8.7 & $\pm$3.2 & \multicolumn{2}{c}{---} & \multicolumn{2}{c}{---} & \multicolumn{2}{c}{---} & \multicolumn{2}{c}{---} & 0 \\
\midrule
Seq.\ memory    & 41.2 & $\pm$5.5 & 16.7 & $\pm$15.6 & 45.6 & $\pm$13.2 & 17.3 & $\pm$10.8 & 49.2 & $\pm$3.3 & 15.0 & $\pm$2.4 & 12{,}065 \\
Batch memory    & 34.4 & $\pm$8.0 & 13.0 & $\pm$8.1 & 49.4 & $\pm$13.7 & 13.7 & $\pm$11.8 & 62.2 & $\pm$11.1 & 22.0 & $\pm$3.6 & 439 \\
ExpeL           & 49.1 & $\pm$6.3 & 12.0 & $\pm$5.2 & 52.2 & $\pm$5.3 & 12.0 & $\pm$1.4 & 55.9 & $\pm$3.6 & 13.3 & $\pm$4.9 & 114 \\
SkillX          & 53.1 & $\pm$4.6 & 21.3 & $\pm$3.6 & 57.5 & $\pm$2.0 & 20.0 & $\pm$3.7 & 55.9 & $\pm$2.5 & 19.3 & $\pm$5.1 & 219 \\
Evo-MedAgent    & 67.8 & $\pm$14.1 & 31.3 & $\pm$21.7 & 69.4 & $\pm$9.2 & 30.7 & $\pm$11.8 & 55.7 & $\pm$11.7 & 26.1 & $\pm$3.5 & 21{,}060 \\
\midrule
\textsc{Grasp} (ours) & 40.1 & $\pm$11.3 & 27.2 &$\pm$24.7 & \bfseries 73.4 & \textbf{\tiny $\pm$23.1} & \bfseries 50.9 & \textbf{\tiny $\pm$20.0} & \bfseries 88.8 & \textbf{\tiny $\pm$5.8} & \bfseries 56.3 & \textbf{\tiny $\pm$14.5} & 411 \\
\bottomrule
\end{tabular}
\label{tab:gated}
\end{table*}

Three explanations for the margin separate here. Validation compute is not one of them, since the matched-compute variants of Section~\ref{sec:results:ablations} spend the full probe budget and discard the verdict, reaching 70.8\% and 67.2\%, and across methods Batch memory shares \textsc{Grasp}'s cadence, gate, probe, and injection field at slightly higher cost, 439 against 411 probe calls, while reaching 62.2\% test and 22.0\% OOD, with the per-episode methods spending 29$\times$ and 51$\times$ \textsc{Grasp}'s probe calls (Table~\ref{tab:gated-compute}). Comparative proposal generation contributes in-domain and not uniformly, gaining Batch memory 12.8 test points from $K{=}1$ to $K{=}4$ and costing Evo-MedAgent 13.7, and only Batch memory moves OOD ($+8.3$), still 34.3 points short. What remains is the gate applied to a bounded, editable library. The same validation signal that buys in-domain fit for a flat memory or an append-only rule list produces generalization only when the edited object is capacity-bounded and supports modification and removal, and within \textsc{Grasp} restricting the writer to \texttt{ADD} costs 8.6 points while dropping the hard budget $R(c) \leq R_0$ costs 7.0.

\subsection{Non-Medical Environments}
\label{sec:results:nonmedical}

Table~\ref{tab:nonmedical} reports test accuracy on four non-clinical AgentBench environments using gpt-oss-120b across three seeds. \textsc{Grasp} produces a large gain on ALFWorld (+28.4) and a substantial gain on WebShop (+20.6), where tasks recur with verifiable structure, a smaller gain on DBBench (+5.0), and no gain beyond seed noise on OS Interaction (+0.9), where the action space is open-ended.

\begin{table}[!h]
\centering
\caption{Non-medical results on four AgentBench~\citep{liu2024agentbench} environments using gpt-oss-120b across three seeds.}
\small
\begin{tabular}{l PP S[table-format=+2.1, print-implicit-plus]}
\toprule
\textbf{Benchmark} & \multicolumn{2}{c}{\textbf{No skills}} & \multicolumn{2}{c}{\textsc{Grasp}} & {\textbf{Gain}} \\
\midrule
ALFWorld          & 23.3 & $\pm$2.9 & 51.7 & $\pm$7.6 & +28.4 \\
WebShop           & 20.7 & $\pm$1.2 & 41.3 & $\pm$8.1 & +20.6 \\
DBBench           & 65.6 & $\pm$3.9 & 70.6 & $\pm$1.0 & +5.0 \\
OS Interaction    & 48.6 & $\pm$2.9 & 49.5 & $\pm$1.7 & +0.9 \\
\bottomrule
\end{tabular}
\label{tab:nonmedical}
\end{table}

\section{Discussion}
\label{sec:discussion}
 
\textsc{Grasp} treats agent self-improvement as a sequence of validated edits to a bounded library of behavioral skills, each candidate gated by a regression-aware probe of previously failing and passing examples. The design responds to a specific failure mode of prior verbal-feedback methods, which accumulate guidance monotonically without checking that each item preserves existing behavior. Our results place the gain in validated skill \textit{editing} rather than in better skill \textit{writing}, since without validation the skill-writer matches the no-skills baseline and the probe's compute does not substitute for its verdict (Section~\ref{sec:results:ablations}). Sequential and Batch memory update more often than \textsc{Grasp} yet regress below the no-skills baseline as context accumulates, illustrating that prompt-based agents forget through dilution rather than parameter drift, which a held-out probe naturally detects.
 
Granting the same gate to the baselines separates the mechanism from the validation step itself (Section~\ref{sec:results:gated}). Held-out validation is not specific to \textsc{Grasp} and lifts every baseline in-domain, none of them out of distribution, and the probe compute does not account for the difference. What produces generalization is the gate applied to a bounded, editable library, in which modification and removal are first-class operations and previously correct behavior is protected by a hard constraint rather than traded off in a scalar objective.
 
Cross-model and cross-benchmark transfer together indicate that the learned skills encode environment-specific procedural knowledge rather than model-specific in-context patterns or general clinical competence. The clearest signal is on held-out task types, where skills written by GPT-5.4 (low) improve every executor's OOD accuracy over what it learns for itself, while the in-domain effect is smaller and can cost a few points on the strongest executors. The direction is asymmetric. A stronger writer improves a weaker executor, but libraries from weaker writers applied to stronger executors are consistently worse than self-training. This asymmetry is specific to \textsc{Grasp}. The same protocol applied to all five baselines leaves every one near or below 35\% OOD after transfer (Appendix~\ref{app:cross-writer-baselines}), indicating that the gate, not writer strength, makes a library transferable. A natural reading is that a stronger writer articulates operational facts about the environment that a weaker executor can follow even when it cannot generate them, a distillation-like effect without parameter updates, and the gain concentrating on held-out task types fits this account. Transfer across benchmarks bounds the same knowledge from the other side. Skills move between MedAgentBench and MedAgentBench-v2, which share a tool-calling convention, but degrade FHIR-AgentBench below its own baseline, because a library written against one convention emits malformed calls under FHIR-AgentBench's Python-tool and free-text interface. What transfers is procedural knowledge tied to an environment and its interface, not clinical FHIR competence in general.
 
The non-medical results delineate where the mechanism applies. \textsc{Grasp} helps most where tasks recur with predictable structure and side effects are directly verifiable (ALFWorld, WebShop), and least where the action space is open-ended and consequences are diffuse, as on OS Interaction, where the gain does not exceed seed noise. DBBench falls between, since only a subset of its query types exhibits the recurring, checkable patterns a skill can target. A separate boundary is set by resource limits, not task structure. On MedAgentBench-v2 the proprietary models fail most often by exhausting the fixed 8-action budget on two paginated-search task types, which no method learned to avoid (Appendix~\ref{app:budget-failures}). The common factor across domains is not subject matter but task structure, recurring failure modes paired with a verifiable signal the probe can use to admit or reject an edit.
 
Two properties make the mechanism practical to deploy. Bounded libraries outperform unbounded memories at a fraction of the inference cost, injecting 6$\times$ to 9$\times$ fewer tokens (Section~\ref{sec:setup:protocol}) and not growing with training length, consistent with prior findings that LLMs degrade under irrelevant context \citep{shi2023large, liu2024lost}. The library is also auditable by design, since each skill is a versioned Markdown document recording which failure mode prompted it, which probe score justified its acceptance, and which earlier skill it replaced (Appendix~\ref{app:learned-skills}), so a reviewer can trace why an agent behaves as it does and revert a specific behavior without retraining. Both matter most where an agent must be inspected and corrected after deployment rather than retrained, as in a clinical setting.

\section{Conclusion}
\label{sec:conclusion}
\textsc{Grasp} converts agent failures into validated edits to a bounded library of behavioral skills, each candidate gated by a regression-aware probe of previously failing and previously passing examples. Against the strongest of five memory-based baselines, \textsc{Grasp} adds 21.0 points on MedAgentBench, and the gain holds across five base models on two FHIR clinical benchmarks. Ablations localize it to the validation gate rather than the skill-writing process, which alone closes none of the gap against the no-skills baseline. Granting the same gate to all five baselines lifts each of them in-domain and none of them out of distribution, placing the gain in the gate applied to a bounded, editable library rather than in held-out validation as such. Frozen libraries written by a stronger model improve weaker executors beyond what they learn for themselves while the reverse does not, an asymmetry that no ungated baseline reproduces and that, with the way skills break across interface boundaries, indicates the libraries encode environment-procedural knowledge rather than model-specific patterns. The same mechanism helps across clinical and non-clinical environments where tasks recur with verifiable structure, and is flat only where the action space is open-ended. Treating agent self-improvement as gated editing of a bounded library, rather than monotonic accumulation of guidance, offers a path toward iterative refinement of LLM agents in environments where procedural reliability matters.

\paragraph{Code and Data Availability.} We release a unified codebase implementing \textsc{Grasp} and all five comparators against a shared agent and evaluation interface, with adapters for all seven benchmarks (\url{https://github.com/jomoll/GRASP}). A \texttt{Task} interface applies \textsc{Grasp} to a new agent and environment, and a \texttt{Method} interface benchmarks a new self-improvement method against all six on any of them. The repository also contains the learned and frozen libraries, all prompts, and per-seed accuracies for every cell of Tables~\ref{tab:main}--\ref{tab:nonmedical}.

\paragraph{Acknowledgments.} This work was supported by Bayern Innovativ (Bavarian State Ministry of Economics) under Grant KK6052501. We also thank François Beaulieu and Paul Vozila of Microsoft Healthcare \& Life Sciences for their support of this project.

\section*{Limitations}
\label{sec:limitations}
\textsc{Grasp} is evaluated on benchmarks that score procedural reliability against ground-truth FHIR state rather than clinical correctness or patient outcomes. MedAgentBench and its variants use curated patient records, a containerized FHIR server, and exact-match scoring against pre-specified answers. Real clinical environments might differ in ways the benchmarks do not capture, including incomplete and contradictory records, large free-text note volumes, non-standard local codes, multi-system integration, and workflows that depend on context outside the EHR. All clinical evaluation is on English-language FHIR data, and generalization to non-English content and to proprietary EHR systems that do not expose FHIR is open. The skills \textsc{Grasp} learns are policies tuned to a benchmark rather than medical expertise, and translating the mechanism to a clinical setting would require clinician review of each skill, prospective comparison against current practice, and monitoring under live workflow conditions, none of which are reported here.
 
The validation step is the source of \textsc{Grasp}'s stability but also its dominant cost. Each batch evaluates the current library plus at most $K$ candidates on a 36-episode probe, contributing 411 agent calls per batch and accounting for the majority of training-time compute. The ``no parameter updates'' framing should not be read as ``no cost''. The cost has moved from gradient updates to LLM inference. In settings where dev-split episodes are expensive, for example live clinical environments or slow tool interfaces, the probe size that delivered our reported results may need to be reduced, with the trade-off between probe size and gate reliability characterized in Appendix~\ref{app:sensitivity}. In their published form the baselines run no probe, so \textsc{Grasp}'s advantage over them could in principle reflect the extra compute the gate consumes rather than the gate's decisions themselves. We separate the two in two ways. Our matched-compute ablations (Section~\ref{sec:results:ablations}) hold the compute fixed within \textsc{Grasp} and vary only the gate, spending the full per-batch probe budget but discarding its verdict when selecting an edit, and accuracy stays at the no-gate level. Section~\ref{sec:results:gated} adds the gate to all five baselines, at which point Batch memory spends slightly more probe compute than \textsc{Grasp} and the per-episode methods spend 29$\times$ and 51$\times$ as much, none of them approaching \textsc{Grasp}'s accuracy. Both comparisons are on MedAgentBench with gpt-oss-120b, and we have not repeated the gated-baseline experiment on the other benchmarks or base models.
 
The cross-model transfer result is asymmetric. Stronger writers produce libraries that improve weaker executors beyond self-training, but the reverse direction is consistently worse than self-training. A frozen library is therefore not fully model-agnostic in practice, and deploying a library written by one model on a different one requires verifying that the new executor is not weaker on the procedural axes the writer assumed. We do not study how to detect such mismatches automatically, and the present results give no guidance on choosing a writer model for a given target executor beyond ``use the strongest available''. The asymmetry itself is specific to \textsc{Grasp}. We ran the cross-writer protocol on all five baselines in the two configurations where \textsc{Grasp} shows its strongest effect (GPT-5.4 to Gemini and to gpt-oss-120b), and none reproduces it (Appendix~\ref{app:cross-writer-baselines}). We do not extend this comparison to the full writer-executor matrix, which we compute only for \textsc{Grasp}.
 
Three of the five base models we evaluate (GPT-5.4, GPT-4.1, Gemini 3.1 Flash Lite) are accessed through provider APIs that may change without notice. Exact model handles and access dates are reported in Appendix~\ref{app:models}, and the proprietary rows of our results tables reflect access conditions specified there.

\bibliography{custom}

\clearpage
\appendix
 
\section{Related Work}
\label{app:related}
 
\paragraph{Verbal-feedback self-improvement.}
Reflexion \citep{shinn2023reflexion} introduced the paradigm in which an agent writes natural-language critiques of its own failed trajectories and conditions on those critiques on subsequent attempts. ExpeL \citep{zhao2024expel} extends this to cross-trajectory rule extraction with bounded AGREE, EDIT, REMOVE, or ADD operations. A critical survey \citep{kamoi2024can} documents that LLMs often cannot reliably correct their own reasoning without external feedback. These approaches accumulate or revise guidance monotonically with no held-out check that a new piece of feedback preserves previously correct behavior. \textsc{Grasp} retains the verbal-feedback paradigm but gates every candidate edit on a regression-aware probe of previously failing and passing examples.
 
\paragraph{Learning from agent failures.}
AutoGuide \citep{fu2024autoguide} creates context-aware guidelines from paired trajectories with different outcomes and retrieves them at inference. 
AgentDebug \citep{zhu2025llm} introduces a modular taxonomy of agent failures and a debugging framework that traces failures to their root cause. ECHO \citep{hu2025sample} adapts hindsight experience replay to LM agents, generating counterfactual successful trajectories from failed ones. Experiential Reflective Learning \citep{allard2026experiential} extracts heuristics from single-attempt trajectories and retrieves them at test time. 
\textsc{Grasp} shares the failure-driven orientation but differs on three axes. The unit of learning is a structured behavioral skill rather than a retrieved trajectory or heuristic, every edit passes through an explicit regression-aware gate rather than relying on accumulation or retrieval, and modification and removal are first-class operations, so the library evolves rather than grows.
 
\paragraph{Skill-library agents.}
Voyager \citep{wang2023voyager} established skill libraries as accumulated behavioral primitives conditioned on at inference. SkillX \citep{wang2026skillx} extends this to general agent tasks by decomposing successful trajectories into named skills retrieved by token overlap, and SAGE \citep{wang2025reinforcement} integrates a skill library into RL-based agent training.
These methods grow their library without validating that each addition preserves prior behavior, and draw skills only from successful trajectories. \textsc{Grasp} retains the structured-library design but is failure-driven, supports modification and removal, gates every edit on probe-based net improvement, and requires no parameter updates.
 
\paragraph{Memory-augmented LLM agents.}
MemGPT \citep{packer2023memgpt} introduced managed long-term memory with explicit page-in and page-out operations, and A-Mem \citep{xu2026mem} organizes agent memories into dynamic interconnected knowledge networks with agent-driven indexing decisions. 
In our experimental comparison, sequential and batch memory \citep{chen2025medagentbench} append correction notes after each failing episode or batch, and Evo-MedAgent \citep{shen2026evo} maintains episodic and semantic memories with top-$k$ retrieval. None of these methods evaluate updates against a held-out probe and the memory grows monotonically over training. In our experiments this leads memory baselines to either regress below the no-skills baseline as context dilutes (Sequential, Batch) or oscillate sharply between epochs (Evo-MedAgent), confirming that update frequency and retrieval are not substitutes for an explicit acceptance criterion.
 
\paragraph{Prompt and program optimization.}
DSPy \citep{khattab2023dspy} treats the natural-language portion of an LLM pipeline as parameters optimized against a development set, using held-out validation to compare candidates. Broader prompt-optimization methods \citep{zhou2022large, yang2024large} cast prompt design as search guided by held-out performance. \textsc{Grasp} shares the held-out validation idea but applies it at a different granularity. The optimized object is a structured library of behavioral skills with add, modify, and remove operations rather than a flat prompt or fixed program template, and the acceptance criterion explicitly bounds regressions on previously correct behavior. This connects \textsc{Grasp} to the broader continual-learning literature on catastrophic forgetting \citep{kirkpatrick2017overcoming}.
 
\paragraph{Clinical agent benchmarks.}
LLM evaluation in clinical settings has historically focused on question answering. Med-PaLM \citep{singhal2023large} established MultiMedQA as a benchmark combining medical licensing exam questions, biomedical literature, and consumer health queries. These benchmarks test medical knowledge but not the procedural reliability required to operate in a clinical environment. MedAgentBench \citep{jiang2025medagentbench} introduced FHIR-based tasks requiring resource retrieval, medication reconciliation, and verified writes. MedAgentBench-v2 \citep{chen2025medagentbench} extends the original with multi-step decision tasks, coordinated writes, and clinical safety protocols. FHIR-AgentBench \citep{lee2025fhir} evaluates structured clinical question answering and tool use on an independent FHIR environment.
 
\paragraph{Structured-environment agents beyond clinical tasks.}
ReAct \citep{yao2022react} introduced the thought-action-observation loop that underlies most modern tool-using agents, and Toolformer \citep{schick2023toolformer} demonstrated that LLMs can learn to invoke external APIs without explicit supervision. AgentBench \citep{liu2024agentbench} provides a standard suite for operating systems, databases, and web interfaces, while Mind2Web \citep{deng2023mind2web} and WebArena \citep{zhou2024webarena} target the web and SWE-bench \citep{jimenez2024swe} targets software engineering. The non-medical experiments in Section~\ref{sec:results:nonmedical} draw from the AgentBench environments.
 
\paragraph{Context degradation under long inputs.}
A consequence of monotonic memory accumulation is that task-relevant guidance is diluted by outdated or redundant entries. \citet{shi2023large} showed that LLMs are sensitive to irrelevant context inserted into their inputs, and \citet{liu2024lost} documented systematic position-based degradation in long-context retrieval. These findings motivate the bounded-library design of \textsc{Grasp}, which keeps the active skill set small by allowing removal and modification rather than only addition.

\section{Method Details}
\label{app:method}
 
\subsection{Per-Batch Algorithm}
\label{app:algorithm}
 
Algorithm~\ref{alg:grasp} gives the full per-batch GRASP update referenced in Section~\ref{sec:method:selection}.
 
\begin{algorithm*}[h]
\caption{GRASP: per-batch skill update}
\label{alg:grasp}
\KwIn{Library $S$, batch $B$ of dev episodes, prior dev runs $\mathcal{H}$, max proposals $K$, probe size $N$}
\KwOut{Updated library $S$}
 
\BlankLine
\tcp{Step 1: Run batch and record outcomes}
$\mathcal{T}_B \leftarrow $ run agent with library $S$ on each episode in $B$ \;
record per-episode correctness in $\mathcal{H}$ \;
$\mathcal{F}_B \leftarrow$ failed traces in $\mathcal{T}_B$ \;
 
\BlankLine
\tcp{Step 2: Group failures and propose}
$L \leftarrow$ \textsc{ClassifyByFailureMode}($\mathcal{F}_B$) \tcp*{LLM call}
$\mathcal{C} \leftarrow$ \textsc{Propose}($\mathcal{F}_B$, $L$, $S$), with $|\mathcal{C}| \leq K$ \tcp*{at most $K$ candidate edits}
 
\BlankLine
\tcp{Step 3: Construct probe from earlier dev runs (never touches val/test)}
$\mathcal{P}_{\mathrm{fail}}, \mathcal{P}_{\mathrm{pass}} \leftarrow$ \textsc{StratifiedSample}($\mathcal{H} \setminus B$, $N/2$, $N/2$) \;
 
\BlankLine
\tcp{Step 4: Baseline run --- re-evaluate current library $S$ on the probe}
$F_0 \leftarrow |\{x \in \mathcal{P}_{\mathrm{fail}} : x \text{ passes under } S\}|$ \;
$R_0 \leftarrow |\{x \in \mathcal{P}_{\mathrm{pass}} : x \text{ fails under } S\}|$ \;
$\mathcal{E}_0 \leftarrow$ samples that errored during baseline run \;
 
\BlankLine
\tcp{Step 5: Score each candidate on the same probe}
\For{$c \in \mathcal{C}$}{
  $S^c \leftarrow$ apply $c$ to a fork of $S$ \;
  $F(c) \leftarrow |\{x \in \mathcal{P}_{\mathrm{fail}} \setminus \mathcal{E}_0 : x \text{ passes under } S^c\}|$ \;
  $R(c) \leftarrow |\{x \in \mathcal{P}_{\mathrm{pass}} \setminus \mathcal{E}_0 : x \text{ fails under } S^c\}|$ \;
  $\mathrm{score}(c) \leftarrow (F(c) - F_0) - (R(c) - R_0)$ \;
}
 
\BlankLine
\tcp{Step 6: Accept the best candidate that satisfies the regression budget}
$\mathcal{C}^\star \leftarrow \{c : \mathrm{score}(c) > 0 \text{ and } R(c) \leq R_0\}$ \;
\If{$\mathcal{C}^\star \neq \emptyset$}{
  $c^\star \leftarrow \arg\max_{c \in \mathcal{C}^\star} \mathrm{score}(c)$ \;
  \If{$R(c^\star) > 0$}{
    $c^\star \leftarrow$ \textsc{ContrastiveRevision}($c^\star$, regressing samples) \tcp*{if revision improves score}
  }
  $S \leftarrow$ apply $c^\star$ to $S$ \;
}
\Return $S$
\end{algorithm*}
 
\subsection{Contrastive Revision}
\label{app:revision}
 
When the highest-scoring candidate $c^\star$ satisfies Equation~\ref{eq:acceptance} but causes at least one regression on the probe, the skill-writer is invoked a second time with the original proposal and the regressing traces, and asked to produce a narrower version that preserves the original fixes while exempting the regression patterns. The revision is evaluated against the same probe and the same $F_0, R_0, \mathcal{E}_0$ as the original candidate. It replaces $c^\star$ only if its adjusted score $(F - F_0) - (R - R_0)$ is strictly higher and it satisfies $R \leq R_0$. If the revision fails either condition, the original candidate is applied unchanged.

\subsection{Invalid-Action Regression Weighting}
\label{app:invalid-action}
 
Regressions from invalid agent actions (malformed tool calls, parse failures, or other agent-side errors the environment rejects) are weighted by $\lambda = 2$ in the score, while the hard budget $R(c) \leq R_0$ uses the unweighted count. The asymmetry reflects cost, not tuning. An invalid action fails the episode at the interface level, and under transfer a single rejected call ends the trajectory, so trading a wrong answer for a malformed call is a worse regression than the raw count indicates. The score penalizes these edits more heavily to discourage them without forbidding them, while the budget stays an unweighted count to keep the constraint interpretable. The result is insensitive to $\lambda$ (Appendix~\ref{app:sensitivity}), since invalid-action regressions are rare on the probe.
 
\subsection{Skill Template}
\label{app:skill-template}
 
GRASP-authored skills follow the structure below, mirroring the Agent Skills SKILL.md convention. Selected examples appear in Appendix~\ref{app:learned-skills}.
 
\begin{verbatim}
---
name: skill_name
description: one-line description
tags: [tag1, tag2]
version: 1
provenance:
  epoch: 0
  update_cycle: 2
  action: ADD
  probe_score: 4
---
 
## Trigger
When [specific condition]...
 
## Rule
You must [behavioral instruction]...
 
## Verification
After [action with side effect], confirm...
 
## Example
Failing: [failing trajectory]
Corrected: [corrected trajectory]
\end{verbatim}
 
\subsection{Skill and Memory Injection}
\label{app:injection}
 
Methods differ in how learned content reaches the agent at inference, in where it is placed, how much is injected, and how often it is refreshed. We hold the placement fixed across methods so that it cannot account for the accuracy differences, and report the remaining differences in injected volume in Appendix Table~\ref{tab:budget}.
On the first decision turn, all methods insert their learned content into the same structured field of the task prompt, the \texttt{behavioral\_skills} field that \textsc{Grasp} uses, falling back to a prepend before the task description only when the prompt is not structured as JSON. \textsc{Grasp} injects its skill library, Sequential and Batch memory inject their accumulated correction-note block, Evo-MedAgent injects its retrieved episodic and semantic memory, ExpeL injects its numbered rule list, and SkillX injects its top-k retrieved skills. Each method keeps its own internal formatting of that content, but the injection point is common to all. On turns after the first, the memory and retrieval methods refresh their content per query in their native manner, appending the current block to the running context, while \textsc{Grasp}'s library is fixed for the episode and is not re-injected.
 
\begin{table*}[!h]
\centering
\caption{LLM call accounting per training batch and library size at end of training, on MedAgentBench with gpt-oss-120b. \emph{Update} calls come from the skill-writer, \emph{probe} calls come from re-running the agent on development-split probe episodes. Each gate decision scores the current library plus the candidates the skill-writer returns for that batch, at most $K$ and 3.68 on average, since proposals are generated per failure-mode group and a batch may yield fewer than $K$ distinct admissible candidates. Probe calls are $N \times (1 + \bar{c}) \times 2.44$ with $N$=36 episodes and 2.44 LLM calls per episode. Table~\ref{tab:gated-compute} reports the same accounting for the gated baselines.}
\small
\setlength{\tabcolsep}{5pt}
\begin{tabular}{l rrr rr l}
\toprule
& \multicolumn{3}{c}{\textbf{Training (per batch)}} & \multicolumn{2}{c}{\textbf{Inference state}} & \\
\cmidrule(lr){2-4} \cmidrule(lr){5-6}
\textbf{Method} & \textbf{Update} & \textbf{Probe} & \textbf{Total} & \textbf{Items} & \textbf{Tokens} & \textbf{Inference context} \\
                & \textbf{calls}  & \textbf{calls} & \textbf{calls} &                &                & \\
\midrule
No skills       &   0    &   0    & 117  &   0  &      0 & none \\
Sequential mem. &  ~28   &   0    & 145  & 332  & 34{,}500 & unbounded \\
Batch mem.      &  ~28   &   0    & 145  & 331  & 36{,}500 & unbounded \\
ExpeL           &  ~14   &   0    & 131  &  20  &  1{,}200 & bounded (cap 20 rules) \\
Evo-MedAgent    &   96   &   0    & 213  & 209  & 53{,}100 & retrieval over growing stores \\
SkillX          &  ~90   &   0    & 207  &   5  & 51{,}600 & retrieval ($k$=5, code skills) \\
\textsc{Grasp} (ours) & 6 & 411 & 534 & 5 & 5{,}600 & capacity-bounded ($\leq$10) \\
\bottomrule
\end{tabular}
 
\label{tab:budget}
\end{table*}
 
\begin{table*}[!h]
\centering
\caption{Base models used in this work. Open-source models are self-hosted, proprietary models are accessed via the listed provider's API. Decoding settings are identical across models within each role (agent, skill-writer), see Section~\ref{sec:setup:protocol}.}
\small
\setlength{\tabcolsep}{3pt}
\begin{tabular}{@{}l l l l l l l@{}}
\toprule
\textbf{Model} & \textbf{Provider} & \textbf{Access} & \textbf{Version / handle} & \textbf{Size} & \textbf{Precision} & \textbf{Access date} \\
\midrule
gpt-oss-120b           & OpenAI    & self-hosted & openai/gpt-oss-120b  & 120B & MXFP4 & 05/2026 \\
DeepSeek V4 Flash      & DeepSeek  & self-hosted & deepseek-ai/DeepSeek-V4-Flash  & 284B (13B active) & FP8 & 05/2026 \\
Gemini 3.1 Flash Lite  & Google    & API         & gemini-3.1-flash-lite-preview & --   & --             & 05/2026 \\
GPT-5.4 (low effort)   & OpenAI    & API         & gpt-5.4-2026-03-05 & --   & --             & 05/2026 \\
GPT-4.1                & OpenAI    & API         & gpt-4.1-2025-04-14 & --   & --             & 05/2026 \\
\bottomrule
\end{tabular}
 
\label{tab:models}
\end{table*}

Holding placement fixed isolates two things that still vary across methods, the learned content itself and the update or selection rule that produces it, which is what the comparison is meant to test. The differences that remain do not favor \textsc{Grasp}. End-of-training injected-token counts in Appendix Table~\ref{tab:budget} show that \textsc{Grasp} injects 6$\times$ to 9$\times$ fewer tokens than the unbounded memory baselines and than SkillX, so the gain cannot come from injecting more text. Batch memory shares \textsc{Grasp}'s injection field and its per-batch update cadence yet regresses below the no-skills baseline (34.4 versus 40.6 on gpt-oss-120b), indicating that injecting content into this field without an acceptance check is not by itself helpful. The no-gate ablation in Section~\ref{sec:results:ablations} removes only the acceptance gate while leaving \textsc{Grasp}'s injection and proposal machinery unchanged, and test accuracy drops by 25.3 points, attributing the gain to the gate rather than to placement.
 
\section{Model Specifications}
\label{app:models}
Table~\ref{tab:models} lists the exact model identifiers, providers, and access details for the five base models used throughout this work. Open-source models (gpt-oss-120b, DeepSeek V4 Flash) are self-hosted using vLLM on an NVIDIA H200 GPU. Proprietary models are accessed via official provider APIs.
All agent execution uses temperature $0.0$ with top-$p = 1.0$. The skill-writer uses temperature $0.7$ with top-$p = 1.0$ for proposal diversity (see Section~\ref{sec:setup:models}). Output is capped at 32{,}768 tokens per call across all methods and models, input prompts use each model's default context window without manual truncation.

\section{Benchmark and Split Details}
\label{app:benchmarks}
 
\begin{table}[!h]
\centering
\caption{Benchmarks and split sizes used in this work. MedAgentBench and MedAgentBench-v2 include an out-of-distribution (OOD) test split formed from held-out task types; the remaining benchmarks do not.}
\small
\setlength{\tabcolsep}{4pt}
\begin{tabular}{l rrrr}
\toprule
\textbf{Benchmark} & \textbf{Dev} & \textbf{Val} & \textbf{Test} & \textbf{OOD} \\
\midrule
\multicolumn{5}{l}{\emph{Clinical (FHIR)}} \\
MedAgentBench          &  96 &  80 &  64 & 60 \\
MedAgentBench-v2       &  96 &  80 &  64 & 60 \\
FHIR-AgentBench        & 120 &  80 &  80 & -- \\
\midrule
\multicolumn{5}{l}{\emph{AgentBench}} \\
DBBench                & 240 & 124 &  60 & -- \\
OS Interaction         &  79 &  56 &  35 & -- \\
ALFWorld               &  26 &  24 &  20 & -- \\
WebShop                & 100 & 50 & 50 & --\\
\bottomrule
\end{tabular}
\label{tab:benchmarks}
\end{table}
 
Table~\ref{tab:benchmarks} reports split sizes for all seven benchmarks. The three clinical benchmarks share a common training and evaluation protocol. The four AgentBench environments are used as exploratory non-clinical probes.
 
\paragraph{MedAgentBench.} Ten clinical task types (300 samples, 30 per type) executed against a live FHIR server hosted in Docker \citep{jiang2025medagentbench}. Tasks include resource retrieval, medication reconciliation, and verified writes. Two task types (tasks 6 and 7) are held out as the OOD split. The remaining eight task types are partitioned per task type into dev (12), val (10), and in-domain test (8) with a fixed seed. Scoring is exact-match against ground-truth answers and FHIR-server state. Reads are evaluated against a fixed server snapshot of pre-existing resources. Writes are evaluated from the trajectory itself, by checking the proposed POST payload against the expected resource. The benchmark's graders never read an agent-written resource back, so no episode mutates the state seen by any later episode and no per-episode server reset is required. This is the benchmark's protocol and not a modification of our harness.
 
\paragraph{MedAgentBench-v2.} A redesigned variant with ten new task types covering imaging follow-up, anticoagulation reconciliation, vital-sign aggregation, catheter-removal protocols, oncology referrals, hypothyroidism management, QTc-prolongation protocols, opioid and naloxone pairing, and vaccination refresh \citep{chen2025medagentbench}. The task types incorporate multi-step decision logic and clinical safety protocols absent from v1. OOD tasks (5 and 7) differ from v1, so the OOD split exercises an independent set of held-out clinical workflows. Split sizes, scoring, and FHIR backend match v1.
 
\paragraph{FHIR-AgentBench.} A read-only structured clinical question-answering benchmark on MIMIC-IV FHIR data hosted on the Google Cloud Healthcare API \citep{lee2025fhir}. Tasks require retrieval and reasoning over FHIR resources but do not write. Evaluation differs from the MedAgentBench variants. Answer correctness is assessed by an LLM judge against a ground-truth answer with prompt-level normalisation, and retrieval is scored by precision and recall against ground-truth resource IDs. The single headline number reported in this paper is LLM-judged answer accuracy. The benchmark is partitioned into dev (120), val (80), and test (80) with no OOD split.
 
\paragraph{AgentBench environments.} Four environments from the AgentBench suite \citep{liu2024agentbench} are evaluated as exploratory non-clinical probes. These are Database (DBBench), Operating System (OS Interaction), ALFWorld~\cite{ALFWorld20}, and WebShop~\cite{webshop2022}. Split sizes are listed in Table~\ref{tab:benchmarks}. DBBench combines real queries from the standard split (partitioned 60/40 by query type) with synthetic aggregation queries. OS Interaction uses worlds 1 to 5 and 7 split 60/40 stratified per world, with world 6 held out. ALFWorld partitions its standard split 60/40 stratified by task type. WebShop uses a round-robin partition of the standard \texttt{webshop-std} index range (instances 0--199) into dev (100), validation (50), and test (50). Scoring uses each environment's native success criterion (task completion, query correctness, or attribute-matching reward).

\section{Extended Results}
\label{app:extended-results}
 \begin{figure}[!h]
\centering
\includegraphics[width=\columnwidth]{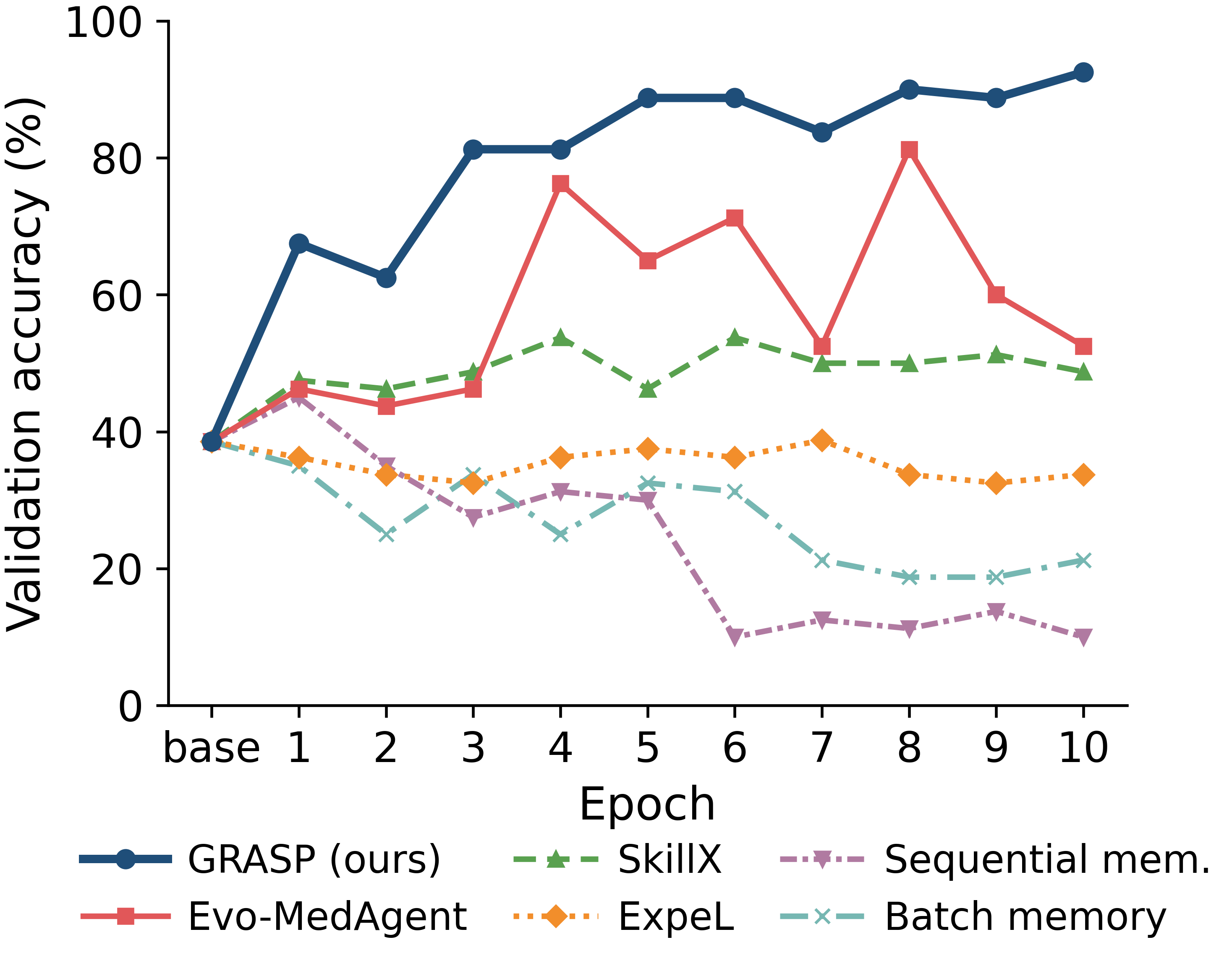}
\caption{Validation accuracy across training epochs on MedAgentBench (gpt-oss-120b). A separate 10-epoch stability run, one seed per method, extended beyond the 5-epoch protocol used for all other results.}\label{fig:learning_curves}
\end{figure}

\subsection{Failure-Mode Classifier: Within-Run Vocabulary Dynamics}
\label{app:classifier-dynamics}
The failure-mode classifier (Section~\ref{sec:method:proposal}) is open-vocabulary. At each batch it assigns each failing trace a label, preferring prior labels and minting new ones only when no prior label fits. We characterize two within-run properties of the resulting vocabulary on MedAgentBench and MedAgentBench-v2 with \texttt{gpt-oss-120b}, across the five seeds used in the main results.

\begin{table}[!h]
\centering
\caption{Cumulative label vocabulary size by batch index. Mean $\pm$ std across 5 seeds.}
\label{tab:classifier-vocab}
\small
\setlength{\tabcolsep}{4pt}
\begin{tabular}{@{}l PP@{}}
\toprule
Batch idx & \multicolumn{2}{c}{MedAgentBench} & \multicolumn{2}{c}{MedAgentBench-v2} \\
\midrule
0 & 11.6 & $\pm$1.4 & 13.0 & $\pm$4.9 \\
2 & 16.4 & $\pm$1.4 & 21.2 & $\pm$6.1 \\
4 & 18.8 & $\pm$2.6 & 24.8 & $\pm$7.8 \\
6 & 19.6 & $\pm$2.9 & 27.8 & $\pm$7.8 \\
8 & 21.0 & $\pm$4.1 & 29.4 & $\pm$8.0 \\
9 & 22.2 & $\pm$4.5 & 30.6 & $\pm$8.2 \\
\bottomrule
\end{tabular}
\end{table}
 
Table~\ref{tab:classifier-vocab} reports the mean cumulative number of distinct labels seen by batch index. The first batch discovers 12 to 13 labels, and the marginal addition per batch drops to 1 or 2 by mid-training. The vocabulary converges rather than expanding indefinitely.
 
\begin{table}[!h]
\centering
\caption{Per-batch trace-level label reuse rate (\%). Fraction of failing traces in batch $b$ whose label appeared in some batch $b' < b$ in the same seed. Mean $\pm$ std across 5 seeds.}
\label{tab:classifier-reuse}
\small
\setlength{\tabcolsep}{4pt}
\begin{tabular}{@{}l PP@{}}
\toprule
Batch idx & \multicolumn{2}{c}{MedAgentBench} & \multicolumn{2}{c}{MedAgentBench-v2} \\
\midrule
1 & 63.9 & $\pm$27.1 & 46.3 & $\pm$20.2 \\
3 & 78.8 & $\pm$18.3 & 68.2 & $\pm$17.3 \\
5 & 80.3 & $\pm$21.1 & 84.3 & $\pm$18.9 \\
7 & 97.5 & $\pm$5.0  & 92.6 & $\pm$4.6 \\
9 & 80.3 & $\pm$20.9 & 88.0 & $\pm$11.0 \\
\bottomrule
\end{tabular}
\end{table}
 
Table~\ref{tab:classifier-reuse} reports the fraction of failing traces per batch whose label was already present in this seed's vocabulary. Reuse climbs from 0\% at batch 0 (by construction) to 80--90\% by batch 4 or 5 and above 90\% by batch 7. After the first epoch, the classifier overwhelmingly reuses prior labels rather than minting new ones.

Together, these two properties show that the classifier converges on a stable label set within a single training run. This is the property the skill-writer needs. By mid-training, failures within a batch are grouped under labels the skill-writer has already seen, allowing edits to target recurring mechanisms rather than novel ones.
 
\subsection{Hyperparameter Sensitivity}
\label{app:sensitivity}
\begin{table}[!h]
\centering
\caption{One-at-a-time sensitivity sweep on MedAgentBench (gpt-oss-120b) around the default configuration $B{=}48$, $N{=}36$, $K{=}4$, $\lambda{=}2$ (marked $^\dagger$). The default row is the same run reused in every block. Three seeds.}
\small
\setlength{\tabcolsep}{6pt}
\begin{tabular}{l l PP}
\toprule
& Setting & \multicolumn{2}{c}{Val$^\star$} & \multicolumn{2}{c}{Test} \\
\midrule
\multirow{3}{*}{$B$ (batch size)}
  & 24            & 87.1 & $\pm$7.3 & 89.1 & $\pm$7.2 \\
  & 48$^\dagger$  & 86.0 & $\pm$4.4 & 88.8 & $\pm$5.8 \\
  & 96            & 83.8 & $\pm$0.0 & 87.0 & $\pm$2.4 \\
\midrule
\multirow{3}{*}{$N$ (probe size)}
  & 16            & 84.6 & $\pm$5.1 & 82.3 & $\pm$3.3 \\
  & 36$^\dagger$  & 86.0 & $\pm$4.4 & 88.8 & $\pm$5.8 \\
  & 72            & 86.2 & $\pm$8.7 & 86.5 & $\pm$10.6 \\
\midrule
\multirow{3}{*}{$K$ (candidates)}
  & 1             & 70.8 & $\pm$25.0 & 73.4 & $\pm$23.1 \\
  & 4$^\dagger$   & 86.0 & $\pm$4.4 & 88.8 & $\pm$5.8 \\
  & 8             & 80.0 & $\pm$7.6 & 84.4 & $\pm$8.3 \\
\midrule
\multirow{3}{*}{$\lambda$ (invalid-action weight)}
  & 1             & 81.7 & $\pm$8.9 & 83.9 & $\pm$10.0 \\
  & 2$^\dagger$   & 86.0 & $\pm$4.4 & 88.8 & $\pm$5.8 \\
  & 4             & 87.1 & $\pm$6.4 & 87.0 & $\pm$7.4 \\
\bottomrule
\end{tabular}
\label{tab:sensitivity}
\end{table}
\noindent Table~\ref{tab:sensitivity} reports a one-at-a-time sensitivity sweep over the parameters governing the per-batch update loop. These are the batch size $B$ (dev episodes per update), the probe size $N$ (episodes used to evaluate candidate libraries), the number of candidate edits $K$ per batch, and the invalid-action regression weight $\lambda$ (Appendix~\ref{app:invalid-action}). Each block varies one parameter around the defaults $B{=}48$, $N{=}36$, $K{=}4$, $\lambda{=}2$ used throughout the paper, with the default run reused as the reference point in every block.
 
\textsc{Grasp} is robust to $B$, $N$, and $\lambda$ but sensitive to $K$. Varying $B$ across $\{24, 48, 96\}$ moves test accuracy by at most 2.1 points (87.0 to 89.1), all within one standard deviation of each other. Probe size $N$ has a floor at 16 (82.3 test) and reaches its full value by 36 (88.8), with no further gain at 72, which the gate-selectivity analysis (Appendix~\ref{app:gate-selectivity}) attributes to the gate already resolving its decisions at $N{=}36$ rather than to a probe too small to discriminate. Varying $\lambda$ across $\{1, 2, 4\}$ leaves in-domain test accuracy in an 84 to 89 band, with all three settings overlapping in standard deviation. $K$ is the exception. At $K{=}1$ test accuracy falls to 73.4 ($\pm$23.1), since a single candidate leaves the gate with a binary accept-or-reject decision and removes the comparative selection that sets \textsc{Grasp} apart from naive verbal-feedback methods, and $K{=}8$ trails $K{=}4$ (84.4 vs 88.8) as more candidates dilute the fixed probe budget. We set $K{=}4$ to give the gate real comparative work without thinning per-candidate evaluation.
 
\begin{table*}[t]
\small
\setlength{\tabcolsep}{3pt}
\centering
\caption{Per-comparison statistics. $\Delta$ is the mean difference (method A minus method B). 95\% CI is a 10{,}000-resample percentile bootstrap on $\Delta$. $p$ is a two-sided unpaired permutation test. $d$ is Cohen's $d$ with pooled SD. Stars mark $p < 0.05$. Daggers ($\dagger$) mark rows where the sample size makes $p < 0.05$ unreachable by construction, the observed $p$ for those rows is the minimum achievable two-sided $p$ at the given $n$.}
\label{tab:sig-tests}
\begin{tabular}{llllrrrrrr}
\toprule
Benchmark & Split & Model & Comparison & A & B & $\Delta$ & 95\% CI & $p$ & $d$ \\
\midrule
MAB     & Test & gpt-oss-120b           & \textsc{Grasp} vs.\ Evo-MedAgent & 88.8 & 67.8 & $+21.0^{\ast}$ & [10.3, 33.8]  & 0.008             & 1.94 \\
MAB     & Test & DeepSeek V4 Flash      & \textsc{Grasp} vs.\ SkillX       & 70.0 & 55.9 & $+14.1^{\ast}$ & [9.4, 18.8]   & 0.008             & 3.24 \\
MAB     & Test & Gemini 3.1 Flash Lite  & \textsc{Grasp} vs.\ Evo-MedAgent & 71.4 & 55.2 & $+16.2$        & [14.1, 18.2]  & $0.100^{\dagger}$ & 8.95 \\
MAB & Test & GPT-4.1        & \textsc{Grasp} vs.\ Batch Memory & 84.9 & 47.4 & $+37.5$ & [27.1, 45.8] & $0.100^{\dagger}$ & 5.53 \\
MAB & Test & GPT-5.4 (low)  & \textsc{Grasp} vs.\ Seq.\ Memory & 85.4 & 47.4 & $+38.0$ & [26.6, 47.4] & $0.100^{\dagger}$ & 4.65 \\
MAB     & OOD  & gpt-oss-120b           & \textsc{Grasp} vs.\ Evo-MedAgent & 56.3 & 31.3 & $+25.0$        & [3.7, 44.0]   & 0.087             & 1.36 \\
MAB-v2  & Test & gpt-oss-120b           & \textsc{Grasp} vs.\ ExpeL        & 76.9 & 67.8 & $+9.1^{\ast}$  & [5.3, 13.1]   & 0.008             & 2.59 \\
MAB-v2  & OOD  & gpt-oss-120b           & \textsc{Grasp} vs.\ SkillX       & 86.0 & 87.1 & $-1.1$         & [$-2.6$, 0.0] & 0.556             & $-0.91$ \\
FHIR-AB & Test & gpt-oss-120b           & \textsc{Grasp} vs.\ no skills    & 58.2 & 51.5 & $+6.8^{\ast}$         & [$+3.7$, 10.0] & 0.008 & 2.36 \\
\bottomrule
\end{tabular}
\end{table*}
 
\subsection{Gate Selectivity}
\label{app:gate-selectivity}
Across five seeds on MedAgentBench with gpt-oss-120b, \textsc{Grasp} applies an edit in 64\% of batches and rejects all four candidates in the other 36\%. Across all candidates it admits 16\%, and on average 1.3 of the 4 per batch clear the regression budget. The accepted candidate's score is a median of 3 and a mean of 4 fixed examples above the acceptance threshold, with a tail reaching 15. These margins are several examples wide on an 18-positive probe, so the gate resolves its decisions well above the one-example floor, which is why a larger probe does not improve accuracy (Table~\ref{tab:sensitivity}).
 
\subsection{Statistical Significance}
\label{app:significance}
Table~\ref{tab:sig-tests} reports per-comparison statistics for the primary results. For each comparison we report the mean difference $\Delta$, a percentile bootstrap 95\% confidence interval on $\Delta$ from 10{,}000 resamples, a two-sided unpaired permutation test (exact when $\binom{n_a+n_b}{n_a} \le 50{,}000$, else Monte Carlo with $10^5$ permutations), and Cohen's $d$ with pooled standard deviation. We treat $\Delta$ and its bootstrap CI as the primary statistical evidence and report permutation $p$-values as supporting detail. Three-seed rows are reported as effect-size evidence rather than as significance claims, and we do not apply multiple-comparisons correction.
 
The seed counts in the main tables, five for open-source models and three for proprietary models and all transfer experiments, imply a minimum achievable two-sided permutation $p$ of $0.008$ at $n_a = n_b = 5$ and $0.10$ at $n_a = n_b = 3$. The latter floor limits the power of permutation testing on three-seed rows but not the interpretation of $\Delta$ and its CI computed on the same data.
 
The primary comparisons show large and reliably positive differences in every cell. On the five-seed open-source comparisons, gpt-oss-120b reaches $\Delta = +21.0$ with 95\% CI $[10.3, 33.8]$ and DeepSeek V4 Flash $\Delta = +14.1$ with CI $[9.4, 18.8]$, both at $p = 0.008$. The three-seed comparisons are of similar or larger magnitude with CIs that exclude zero. Gemini 3.1 Flash Lite reaches $\Delta = +16.2$ with CI $[14.1, 18.2]$, GPT-4.1 $\Delta = +37.5$ with CI $[27.1, 45.8]$, and GPT-5.4 (low) $\Delta = +38.0$ with CI $[26.6, 47.4]$, each with the permutation $p$ capped at $0.10$ by the three-seed protocol. The corresponding Cohen's $d$ values appear in Table~\ref{tab:sig-tests}, where the Gemini-cell $d$ of $8.95$ is inflated by an unusually low pooled standard deviation in that row and should not be read as an estimate of typical effect magnitude. The MedAgentBench-v2 in-domain comparison on gpt-oss-120b is also strong, $\Delta = +9.1$ with CI $[5.3, 13.1]$ at $p = 0.008$.
 
The MedAgentBench OOD comparison on gpt-oss-120b shows $\Delta = +25.0$ with a bootstrap 95\% CI of $[3.7, 44.0]$ that excludes zero. Evo-MedAgent's seed-to-seed standard deviation of $21.7$ points on this split widens the permutation null to $p = 0.087$ but does not shift the underlying gap.
 
On MedAgentBench-v2 OOD, \textsc{Grasp} scores $1.1$ points below SkillX with a bootstrap CI that includes zero, and we make no claim of superiority on this split. The cross-model transfer experiments use three seeds per cell, so we do not assign $p$-values to individual cells and interpret those effects through $\Delta$ and the standard deviations reported in the main tables. Cross-benchmark transfer and the FHIR-AgentBench in-domain comparison use five seeds per cell, with bootstrap CIs reported alongside the point estimates.

\subsection{Cross-Writer Transfer Across Methods}
\label{app:cross-writer-baselines}
 
Table~\ref{tab:cross-writer-baselines} tests whether the cross-model asymmetry in Section~\ref{sec:results:cross-model} reflects \textsc{Grasp}'s gate or any structured guidance from a stronger writer, by running the cross-writer protocol on all five baselines. For each method, GPT-5.4 (low) acts as both agent and writer during training, the resulting library is frozen, and a weaker executor (Gemini 3.1 Flash Lite or gpt-oss-120b) runs at test time.
 
\begin{table}[!h]
  \centering
\caption{Cross-writer transfer on MedAgentBench across three seeds. GPT-5.4 $\rightarrow$ Gemini applies a GPT-5.4-trained library to Gemini 3.1 Flash Lite at test time. GPT-5.4 $\rightarrow$ gpt-oss applies the same library to gpt-oss-120b.}
  \label{tab:cross-writer-baselines}
  \small
  \setlength{\tabcolsep}{4pt}
  \begin{tabular}{@{}l l PP@{}}
  \toprule
  Method & Setting & \multicolumn{2}{c}{ID Test} & \multicolumn{2}{c}{OOD} \\
  \midrule
  \multirow{5}{*}{\textsc{Grasp}}
      & GPT-5.4 self-train            & 85.4 & $\pm$10.4 & 80.6 & $\pm$6.7 \\
      & Gemini self-train             & 71.4 & $\pm$1.8  & 41.7 & $\pm$20.9 \\
      & GPT-5.4 $\rightarrow$ Gemini  & \bfseries 76.6 & \textbf{\tiny $\pm$10.9} & \bfseries 71.1 & \textbf{\tiny $\pm$1.9} \\
      & gpt-oss self-train            & 88.8 & $\pm$5.8  & 56.3 & $\pm$14.5 \\
      & GPT-5.4 $\rightarrow$ gpt-oss & \bfseries 76.0 & \textbf{\tiny $\pm$7.7} & \bfseries 77.8 & \textbf{\tiny $\pm$1.0} \\
    \midrule
    \multirow{5}{*}{SkillX}
      & GPT-5.4 self-train            & 44.8 & $\pm$0.9  &  3.3 & $\pm$2.9 \\
      & Gemini self-train             & 54.2 & $\pm$3.6  & 38.3 & $\pm$1.7 \\
      & GPT-5.4 $\rightarrow$ Gemini  & 56.8 & $\pm$0.9  & 35.0 & $\pm$1.7 \\
      & gpt-oss self-train            & 53.1 & $\pm$4.6  & 21.3 & $\pm$3.6 \\
      & GPT-5.4 $\rightarrow$ gpt-oss & 46.4 & $\pm$8.6  & 17.2 & $\pm$3.8 \\
    \midrule
    \multirow{5}{*}{Evo-MedAgent}
      & GPT-5.4 self-train            & 43.8 & $\pm$4.1  &  1.7 & $\pm$1.7 \\
      & Gemini self-train             & 55.2 & $\pm$1.8  & 15.6 & $\pm$2.5 \\
      & GPT-5.4 $\rightarrow$ Gemini  & 54.2 & $\pm$1.8  & 18.3 & $\pm$6.0 \\
      & gpt-oss self-train            & 67.8 & $\pm$14.1 & 31.3 & $\pm$21.7 \\
      & GPT-5.4 $\rightarrow$ gpt-oss & 44.3 & $\pm$3.9  &  8.3 & $\pm$3.3 \\
    \midrule
    \multirow{5}{*}{Seq.\ Memory}
      & GPT-5.4 self-train            & 47.4 & $\pm$5.0  & 15.6 & $\pm$25.5 \\
      & Gemini self-train             & 53.9 & $\pm$3.3  &  2.5 & $\pm$1.2 \\
      & GPT-5.4 $\rightarrow$ Gemini  & 54.2 & $\pm$3.3  & 11.1 & $\pm$7.5 \\
      & gpt-oss self-train            & 41.2 & $\pm$5.5  & 16.7 & $\pm$15.6 \\
      & GPT-5.4 $\rightarrow$ gpt-oss & 44.8 & $\pm$11.9 & 13.3 & $\pm$10.4 \\
    \midrule
    \multirow{5}{*}{Batch Memory}
      & GPT-5.4 self-train            & 46.4 & $\pm$0.9  & 16.7 & $\pm$3.3 \\
      & Gemini self-train             & 32.0 & $\pm$25.4 &  2.5 & $\pm$1.2 \\
      & GPT-5.4 $\rightarrow$ Gemini  & 49.5 & $\pm$5.9  & 12.2 & $\pm$7.9 \\
      & gpt-oss self-train            & 34.4 & $\pm$8.0  & 13.0 & $\pm$8.1 \\
      & GPT-5.4 $\rightarrow$ gpt-oss & 37.0 & $\pm$11.7 & 14.4 & $\pm$8.6 \\
    \midrule
    \multirow{5}{*}{ExpeL}
      & GPT-5.4 self-train            & 43.8 & $\pm$1.6  & 11.7 & $\pm$3.3 \\
      & Gemini self-train             & 53.6 & $\pm$1.8  & 17.2 & $\pm$2.5 \\
      & GPT-5.4 $\rightarrow$ Gemini  & 52.6 & $\pm$0.9  & 18.9 & $\pm$4.2 \\
      & gpt-oss self-train            & 49.1 & $\pm$6.3  & 12.0 & $\pm$5.2 \\
      & GPT-5.4 $\rightarrow$ gpt-oss & 39.6 & $\pm$3.9  & 12.2 & $\pm$3.5 \\
    \bottomrule
    \end{tabular}
    \end{table}
    
The asymmetry does not replicate on any baseline. \textsc{Grasp} is the only method whose GPT-5.4 library generalizes OOD when transferred, reaching $71.1$ on Gemini and $77.8$ on gpt-oss, while no baseline's transferred library exceeds $35.0$ (SkillX$\rightarrow$Gemini) and most stay below $20$ on both executors. The same frozen GPT-5.4 library reaches comparable OOD on two different executors, and in both cases the transfer exceeds the target's own self-trained library ($41.7$ for Gemini, $56.3$ for gpt-oss), so the gains track the source library rather than the executor. Writer strength alone does not explain this, since the baselines' GPT-5.4 libraries are weak even at the source ($1.7$ to $16.7$ OOD under self-training), leaving little procedural knowledge for transfer to carry. gpt-oss is a capable executor, yet baseline transfers to it stay near or below its self-train OOD rather than benefiting from the GPT-5.4 source. What survives transfer to a weaker executor is the procedural knowledge the gate selects for, not writer strength alone.
 
\subsection{Action-Budget Failures on MedAgentBench-v2}
\label{app:budget-failures}
 
\begin{table}[!h]
\centering
\caption{Action-budget exhaustion on MedAgentBench-v2 for GPT-5.4 (low), as averaged limit-hit episodes per run out of 64 samples. \emph{Base} is the bare model under each method's evaluation conditions, \emph{Adapted} is the method's best-validation checkpoint.}
\label{tab:budget-failures}
\small
\begin{tabular}{@{}l cc r@{}}
\toprule
Method & Base & Adapted & $\Delta$ \\
\midrule
Evo-MedAgent   & 5.7 & 5.0 & $-0.7$ \\
Seq.\ Memory   & 7.0 & 7.0 & $0.0$ \\
ExpeL          & 7.0 & 7.7 & $+0.7$ \\
Batch Memory   & 7.0 & 9.0 & $+2.0$ \\
SkillX         & 6.0 & 9.3 & $+3.3$ \\
\textsc{Grasp} & 6.7 & 8.7 & $+2.0$ \\
\bottomrule
\end{tabular}
\end{table}
 \begin{table}[!h]
\centering
\caption{Training-time LLM calls per batch under the gated $K{=}4$ configuration on MedAgentBench with gpt-oss-120b. \emph{Cadence} is the method's native update frequency, at which the gate fires. Probe cost scales with cadence and with candidate diversity, since ExpeL's \texttt{AGREE}/\texttt{EDIT}/\texttt{REMOVE}/\texttt{ADD} operation sets and SkillX's extracted skills collapse to few distinct library states at $K{=}4$.}
\small
\setlength{\tabcolsep}{4pt}
\begin{tabular}{l rrr l}
\toprule
\textbf{Method} & \textbf{Update} & \textbf{Probe} & \textbf{Total} & \textbf{Cadence} \\
\midrule
Seq.\ memory  & 112 & 12{,}065 & 12{,}294 & per episode \\
Evo-MedAgent  & 384 & 21{,}060 & 21{,}561 & per episode \\
Batch memory  & 112 & 439 & 668 & per batch \\
ExpeL         &  56 & 114 & 287 & per batch \\
SkillX        & 360 & 219 & 696 & per batch \\
\textsc{Grasp} (ours) & 6 & 411 & 534 & per batch \\
\bottomrule
\end{tabular}
\label{tab:gated-compute}
\end{table}

On MedAgentBench-v2, the proprietary models GPT-5.4 (low) and GPT-4.1 most often fail by exhausting the fixed 8-action budget rather than by producing an incorrect answer. The effect concentrates on two task types involving paginated FHIR search (tasks~8 and~10), where the agent spends actions paging through search results and reaches the budget before completing the task. These two tasks reach the action limit on roughly 37--40\% of attempts for GPT-5.4 (low), independent of the adaptation method.
 
Table~\ref{tab:budget-failures} reports the average number of limit-hit episodes per run on MedAgentBench-v2 for GPT-5.4 (low), out of 64 evaluation samples, comparing the bare base model under each method's evaluation conditions against the method's best-validation checkpoint. The base-model column varies only by stochastic decoding, and its spread (5.7--7.0) is within Poisson noise for events at this rate. After adaptation, three methods stay at the base rate within noise and three sit modestly above it, but this split does not correspond to whether a method injects skills. Batch Memory, which injects only a flat memory block, shows the same increase as the skill-based methods, while Sequential Memory, which injects the same kind of memory, does not. With two to three runs per method these differences of one to two additional limit hits are directional rather than statistically resolved. None of the methods reduces the underlying failure, which is set by the base model's behavior on paginated search rather than by any adaptation.

\section{Ground-Truth Audit of the Learned Libraries}
\label{app:audit}
 
The acceptance gate consumes only per-example pass/fail outcomes and therefore cannot encode task answers. The failure classifier and the skill-writer do see expected answers, so we audited all 106 learned skill libraries against ground truth for every split (dev, validation, in-domain test, OOD), matching by skill section.
 
\paragraph{Result.} No held-out-split ground-truth value appears in any load-bearing section (\texttt{Trigger}, \texttt{Rule}, \texttt{Verification}) of any library. Thirty episode-specific values do appear under those headers, all of dev-split origin, and consist of dev-split timestamps and a single MRN. Without exception each sits inside a demonstration context, a fenced code block, a correct/wrong contrast, or an inline \texttt{e.g.} format illustration, and none appears as a free-standing rule statement. The only collisions with hidden splits are four low-entropy domain constants, a calcium unit-conversion factor and a potassium reference range.
 
\paragraph{Interpretation.} The values the audit flags belong to skills such as \texttt{iso8601\_datetime\_with\_timezone}, which requires date-times to carry a timezone offset, and \texttt{require\_fhir\_query\_before\_answer}, which fixes the shape of the terminating call. These are facts about the environment's I/O contract, which is what the paper claims skills encode, and the same holds for the calcium and potassium constants. Empirically, memorized dev answers cannot reach held-out task types, and this is where \textsc{Grasp}'s margins are largest ($+25.0$ OOD on gpt-oss-120b, 80.6 against 16.7 on GPT-5.4).
 
\section{Selected Skills Learned by \textsc{Grasp}}
\label{app:learned-skills}
 
The four skills below were selected by the authors, not sampled at random, to illustrate the format of \textsc{Grasp}-authored skills across a range of conditions. They span four benchmarks (ALFWorld, DBBench, MedAgentBench, MedAgentBench-v2), two authoring models (\texttt{gpt-oss-120b} and \texttt{GPT-5.4} low effort), and two edit types (ADD and MODIFY). The selection is intended to show what \textsc{Grasp} can produce, not what it typically produces. We do not claim these four are representative of average skill quality, length, or specificity across runs. We include these particular four because they make the structural elements of a skill (trigger, rule, verification, contrastive example, provenance) inspectable across the procedural domains the paper evaluates.
 
The clinical skills (Opioid–Naloxone Verification and Patient Identifier to FHIR Reference Resolution) encode procedures for operating MedAgentBench and MedAgentBench-v2 as software systems. They contain medical-domain vocabulary (drug names, RxNorm-style identifiers, FHIR resource types) because the benchmarks do, but the operational content is about the benchmark's API, identifier scheme, and scoring criteria, not about real-world clinical care. The RxNorm codes shown in Opioid–Naloxone Verification, for example, are the codes used by MedAgentBench-v2's FHIR fixture, the skill captures the benchmark's coding scheme rather than the production RxNorm vocabulary. Reading these skills as drafts of clinical decision rules would misidentify what \textsc{Grasp} has learned. They are policies for completing benchmark episodes under exact-match scoring. The clinical scope of these experiments is discussed further in the Limitations section.
 
\clearpage
\onecolumn
\subsection*{Enforce Cleaning Before Placement \\ \normalfont\small (non-clinical, recurring workflow constraint; \texttt{gpt-oss-120b}, AgentBench / ALFWorld; ADD)}
 
\begin{lstlisting}[basicstyle=\ttfamily\footnotesize,breaklines=true,
                   frame=single,framesep=4pt,xleftmargin=4pt,xrightmargin=4pt]
---
name: enforce_clean_before_placement
description: Clean target object before placing into final receptacle
tags: [alfworld, workflow-order]
version: 1
provenance:
  epoch: 3
  action: ADD
---
 
## Pattern Description
 
The target object must never be placed into its final receptacle before
it has been cleaned. The constraint applies to any task whose
instruction mentions "clean" (e.g. "pick_clean_then_place", "clean the
X before placing") or whose query type is pick_clean_then_place.
Enforcing a cleaning action immediately after pickup and before any
placement guarantees that the environment state satisfies the
cleanliness predicate; otherwise the episode silently fails.
 
## Triggers
 
- Instruction contains the word "clean" (case-insensitive) or the
  phrase "clean the <object>".
- Query type is pick_clean_then_place.
- A `take <obj> from <loc>` action has just executed and no prior
  `clean` observation exists for the carried object.
 
## Example
 
### Wrong trajectory
 
Think: The task is "pick_clean_then_place-Spatula-None-Drawer-10".
Act:   take spatula 1 from drawer 2
Obs:   You are now carrying spatula 1.
Think: I have the spatula, place it.
Act:   put spatula 1 in/on drawer 1
Obs:   Spatula 1 is now in drawer 1.
       [Episode ends; goal NOT satisfied.]
 
### Correct trajectory
 
Think: Instruction mentions "clean" -- enforce cleaning before
       placement.
Act:   take spatula 1 from drawer 2
Obs:   You are now carrying spatula 1.
Think: Before placing, I must clean the spatula.
Act:   clean spatula 1 with sinkbasin 1
Obs:   Spatula 1 is now clean.
Act:   put spatula 1 in/on drawer 1
Obs:   Spatula 1 is now in drawer 1.
       [Goal satisfied.]
 
## Pitfalls
 
- Do not clean after placement.
- If the object is already clean, skip the redundant action.
- Choose a valid cleaning device (sinkbasin, dishwasher); other
  receptacles raise an invalid-action error.
- Do not trigger on tasks that do not mention cleaning; this wastes
  the step budget.
\end{lstlisting}
 
  \clearpage
 
\subsection*{Parse Percentage Strings for Aggregation \\ \normalfont\small (non-clinical, low-gain benchmark; \texttt{gpt-oss-120b}, AgentBench / DBBench; ADD)}
 
\begin{lstlisting}[basicstyle=\ttfamily\footnotesize,breaklines=true,
                   frame=single,framesep=4pt,xleftmargin=4pt,xrightmargin=4pt]
---
name: parse_percentage_strings_for_aggregation
description: Strip percent signs and normalise commas before numeric ops
tags: [dbbench, sql, locale-aware]
version: 1
provenance:
  epoch: 2
  action: ADD
---
 
## Pattern Description
 
When a task asks for a numeric aggregation (SUM, AVG, MIN, MAX) and
the table stores percentages as text with a comma decimal separator
and a trailing % sign (e.g. "23,3%"), the raw string cannot be
compared or summed. The agent must strip the % character and replace
the comma with a dot before casting to a numeric type. The same rule
applies whenever a WHERE clause filters on such a column.
 
## Triggers
 
- Instruction contains a percentage written with a comma and trailing
  % (e.g. "23,3%").
- Schema includes a column whose name contains % (e.g.
  `Foreign nationals in %`).
- Query type is an aggregation and the task mentions a percentage
  filter.
 
## Recommended Patterns
 
### Filter on a percentage column
 
-- WRONG
SELECT SUM(pop) FROM tbl
WHERE `Foreign nationals in %` = '23,3%';
 
-- CORRECT
SELECT SUM(pop) AS total
FROM   tbl
WHERE  CAST(REPLACE(REPLACE(`Foreign nationals in %`,
                            '%', ''), ',', '.')
            AS DECIMAL(10,2)) = 23.3;
 
### Aggregate a numeric column that contains commas
 
-- WRONG (keeps commas)
SELECT SUM(`Population`) FROM tbl;
 
-- CORRECT
SELECT SUM(CAST(REPLACE(`Population`, ',', '') AS UNSIGNED))
       AS total_population
FROM   tbl;
 
## Failure Indicators
 
- Observation shows 0 rows or NULL when a non-zero result is expected.
- SQL error: "Incorrect decimal value: '23,3%'".
- Final answer is a string that still includes the % sign.
\end{lstlisting}
 
  \clearpage
 
\subsection*{Opioid--Naloxone Verification \\ \normalfont\small (benchmark-procedural with medical vocabulary; \texttt{gpt-oss-120b}, MedAgentBench-v2; MODIFY)}
 
\begin{lstlisting}[basicstyle=\ttfamily\footnotesize,breaklines=true,
                   frame=single,framesep=4pt,xleftmargin=4pt,xrightmargin=4pt]
---
name: opioid_naloxone_verification
description: Ensure active opioid orders have matching naloxone prescription
tags: [medagentbench-v2, medication-safety, rxnorm]
version: 2
provenance:
  epoch: 2
  action: MODIFY
---
 
## Pattern Description
 
Every active opioid analgesic order must have a matching naloxone
prescription. The skill encodes a four-step procedure: locate opioids
by explicit RxNorm codes, confirm status == active, search for a
naloxone counterpart, and only if absent, create a naloxone
MedicationRequest.
 
## Common Failure Patterns
 
- Missing code filter, treating the whole bundle as if it contained
  only opioids.
- Ignoring status, counting completed or entered-in-error orders as
  active.
- Incorrect naloxone code, matching free-text instead of an exact
  CPT/RxNorm code.
- Empty bundle != no opioids; a bundle may carry non-opioid meds and
  each entry must be checked.
 
## Procedure
 
### Step 1. Identify active opioid orders
 
GET /MedicationRequest?patient={patientId}
 
opioid_codes = [
  "860975",   # hydromorphone (RxNorm)
  "860976",   # oxycodone
  "860977",   # fentanyl
  "860978",   # hydrocodone
  "860979",   # morphine
]
# Keep entries whose medicationCodeableConcept.coding[].code is in
# opioid_codes AND whose status in {active, draft, on-hold}.
 
### Step 2. Search the same bundle for naloxone (RxNorm 860980)
 
If absent and the opioid list is non-empty, POST:
 
POST /MedicationRequest
{
  "resourceType": "MedicationRequest",
  "status": "active",
  "intent": "order",
  "medicationCodeableConcept": {
    "coding": [{
      "system": "http://www.nlm.nih.gov/research/umls/rxnorm",
      "code":   "860980",
      "display":"Naloxone"
    }]
  },
  "subject": { "reference": "Patient/{patientId}" },
  "authoredOn": "{today_iso}"
}
 
### Step 3. Report exactly one of three outcomes
 
FINISH(["No active opioid analgesic orders found ..."])      # case A
FINISH(["Active opioid orders have matching naloxone ..."])  # case B
FINISH(["Naloxone order created for patient ..."])           # case C
 
## Success Indicators
 
- Filters coding[].code against the opioid list and checks
  status == active before deciding.
- POSTs naloxone with the exact RxNorm code only when truly missing.
- FINISH message accurately reflects whether a naloxone order was
  created or not.
\end{lstlisting}
 
  \vspace{1em}
 
\subsection*{Patient Identifier to FHIR Reference Resolution \\ \normalfont\small (benchmark-procedural, cross-model writer contrast; \texttt{GPT-5.4} low effort, MedAgentBench; MODIFY)}
 
\begin{lstlisting}[basicstyle=\ttfamily\footnotesize,breaklines=true,
                   frame=single,framesep=4pt,xleftmargin=4pt,xrightmargin=4pt]
---
name: patient_identifier_to_fhir_reference_resolution
description: Resolve MRN to Patient.id before any dependent FHIR request
tags: [medagentbench, fhir, identifier-resolution]
version: 3
provenance:
  epoch: 4
  action: MODIFY
---
 
## Pattern Description
 
When a task supplies an MRN it must be treated as a search key, not a
guaranteed downstream resource reference. The agent must first issue
GET /Patient?identifier=..., read Patient.id from the response, and
only then issue dependent searches or writes that need patient= or
subject.reference.
 
The skill prevents two distinct failure modes: (i) firing dependent
requests before the patient lookup response arrives, and (ii) reusing
the raw input MRN, stripped digits, placeholders, or template text
where a resolved Patient.id is required.
 
## Common Failure Patterns
 
- GET /Observation?patient=S0789363&code=A1C without first parsing
  entry[0].resource.id.
- GET /Observation?patient=6550627&code=A1C after stripping the
  leading 'S'.
- "subject":{"reference":"Patient/"} or "Patient/UNKNOWN" in POST
  bodies.
- Concatenating actions on one line, e.g. GET /Patient... GET
  /Observation... FINISH(...) before the first response arrives.
- Repeating the patient lookup multiple times after one successful
  match instead of caching the resolved id.
 
## Recommended Pattern
 
# 1. Resolve once.
GET /fhir/Patient?identifier={input_identifier}
 
# 2. Read the bundle.
resolved_id = bundle.entry[0].resource.id
 
# 3. Use that resolved id in every later step:
#    - searches:  patient={resolved_id}
#    - writes:    "subject": {"reference": "Patient/{resolved_id}"}
 
## Example
 
### Correct trajectory
 
GET /fhir/Patient?identifier=S0789363
   -> entry[0].resource.id = "S0789363"
GET /fhir/Observation?patient=S0789363&code=A1C
POST /fhir/ServiceRequest  {
  ...,
  "subject":{"reference":"Patient/S0789363"}
}
FINISH([5.2, "2022-08-09T15:33:00+00:00"])
 
### Wrong trajectory
 
GET /fhir/Observation?patient=6550627&code=A1C   # stripped 'S'
POST /fhir/ServiceRequest {
  "subject":{"reference":"Patient/UNKNOWN"}
}
FINISH([-1])                                     # never resolved
 
## Success Indicators
 
- Bundle is read before any dependent action.
- Exactly one patient lookup per task.
- Every later patient= parameter and Patient/{id} reference matches
  entry[0].resource.id; no transformed, blank, or placeholder
  references appear.
\end{lstlisting}
\end{document}